\documentclass[twocolumn,10pt]{IEEEtran}%

\usepackage{amssymb}
\usepackage{lipsum}
\usepackage[intlimits]{amsmath}
\usepackage{amsfonts}
\usepackage{amsthm}
\usepackage{mathptmx}
\usepackage{graphicx}
\usepackage{makeidx}
\usepackage{fancyhdr}
\usepackage{lettrine}
\usepackage{mathpazo}

\usepackage{avant}
\usepackage{microtype}
\usepackage{color}
\usepackage{setspace}
\usepackage{enumitem}

\usepackage{listings}
\usepackage{algorithm}
\usepackage{algorithmicx}
\usepackage{algpseudocode}

\algnewcommand\algorithmicoutput{\textbf{Output:}} 
\algnewcommand\Output{\item[\algorithmicoutput]}
\algnewcommand\algorithmicinput{\textbf{Input:}} 
\algnewcommand\Input{\item[\algorithmicinput]}

\usepackage[table]{xcolor}

\usepackage{framed}
\definecolor{shadecolor}{gray}{0.9}

\newcounter{theExample}
\newcounter{theRemark}
\newenvironment{Remark}{
	\par\smallskip\refstepcounter{theRemark}%
	\noindent%
	\textbf{Remark~\arabic{theRemark}}:~%
	\ignorespaces%
}{
	\par\smallskip
}

\newtheorem{exmplA}{Example}
\newenvironment{exmpl}
{\begin{shaded}\begin{exmplA}\small\hangindent=7mm \addtolength{\parindent}{7mm}}
		{\end{exmplA}\ignorespacesafterend\end{shaded}\ignorespacesafterend}
	
\begin{document}
	\title{Understanding the Basis of Graph Convolutional Neural Networks via an Intuitive Matched Filtering Approach}
	
	\author{
		Ljubi\v{s}a~Stankovi\'{c},~\IEEEmembership{Fellow,~IEEE,}
		Danilo Mandic,~\IEEEmembership{Fellow,~IEEE,}
		\thanks{
			L. Stankovi\'{c} is with the  University of
			Montenegro, Podgorica, Montenegro. D. Mandic is with the Imperial College London, London, United Kingdom.
			Contact e-mail: ljubisa@ucg.ac.me}
	}
	\maketitle

	\setcounter{tocdepth}{3}

	\begin{abstract}
Graph Convolutional Neural Networks (GCNN) are becoming a preferred model for data processing on irregular domains, yet their analysis and principles of operation are rarely examined due to  the black box nature of NNs. To this end, we revisit the operation of GCNNs and show that their convolution layers effectively perform matched filtering of input data with the chosen patterns (features). This allows us to provide a unifying account of GCNNs through a matched filter perspective, whereby the nonlinear ReLU and max-pooling layers are also discussed within the matched filtering framework. This is followed by a step-by-step guide on information propagation and learning in GCNNs. It is also shown that standard CNNs and fully connected NNs can be obtained as a special case of GCNNs. A carefully chosen numerical example guides the reader through the various steps of GCNN operation and learning both visually and numerically.
	\end{abstract}

\section{Introduction}

Over the last decade, neural networks (NN) have regained popularity, mainly due to advances in deep learning (DL) and convolutional neural networks (CNN). However, NN based analysis and processing of signals and images of a large sample size remains a challenge as every input sample/pixel needs to be associated with one neuron at the input layer. In addition, for meaningful analysis a neural network requires at least one hidden layer, so that even for a simple fully connected (FC) hidden layer, the number of weights increases -- the so called Curse of Dimensionality -- thus making the dimensionality an inevitable bottleneck.  In practical applications, this issue is partially mitigated by exploiting the fact that most physical data  exhibit a smooth nature, with the neighboring signal samples or image pixels exhibiting similarity of some sort. This in turn allows us to exploit local information in the form of  features (patterns), which describe the analyzed signals/images. In this way, our task then boils down to searching for specific localized features (patterns) in data, instead of  standard brute force approaches. 
Another advantage of operating in the feature space is that this resolves the problem of position change of the patterns in data due, for example, translation. Namely, if a certain data feature changes its position, then a standard sample-wise approach would assume a complete change in samples, while a feature-wise approach will look for specific shapes anywhere in the signal, even if delayed. 

This rationale underpins the operation of convolutional neural networks (CNNs), which effectively perform  search for features in the analyzed signal, such that these features are invariant to  position change \cite{kuo2016understanding,mandic2001recurrent,kiranyaz20211d,kuo2017cnn}. More specifically, the window used in the convolution within CNNs (referred to as the \textit{convolution filter or convolution kernel}) is designed to recognize a feature within the signal in hand that is matched to the  kernel waveform form. In doing so, feature matching is performed over the whole signal, akin to a mathematical lens in search of specific forms.

When it comes to irregular domains, graph neural networks (GNNs) is an area which considers graphs in conjunction with neural networks. It benefits from the the universal approximation property of neural networks, pattern matching inherent to  convolutional neural networks, and the ability of graphs to capture irregular relations among the analyzed data samples.  The beginning of graph neural networks can be traced  back to  a decade ago \cite{gori2005new, scarselli2008graph, micheli2009neural,niepert2016learning}, while recent developments have been centered around graph convolutional networks (GCNNs). Benefiting from their intrinsic graph structure, GCNNs account for complex implicit coupling among data and information aggregation when processing (or filtering) samples associated with each vertex.

Recent literature on GCNNs \cite{zhou2018graph, wu2020comprehensive} typically considers the learning aspects, while assuming the stationarity (via shift invariance of convolution operations) and  compositionality (via downsampling or pooling operations) of CNNs.

The goal of this note is to revisit the operation of GCNNs, starting from basic principles. This is achieved by illuminating that the convolution operation in classical CNN rests upon the classical concept of matched filter. This perspective is verified over an intuitive example and then generalized to data on graphs and GCNNs. The matched filter platform is also used to explain most of the functionality of GCNNs, including nonlinear activation functions and max-polling. Our approach is supported by a step-by-step worked example based on a signal on a graph.

\section{Convolution -- Matched filter} 

\subsection{Time domain convolution} 

The use of a convolutional window has become a de facto standard in CNNs, yet the key open question of how we can justify that the convolution is an appropriate operation for detecting features in a signal remains largely unanswered -- a subject of this note. To this end, we draw inspiration from  matched filter theory, whereby convolution of the considered signal with the feature of interest serves to confirm the existence and location of the feature at hand within the analyzed signal. Recall that the output of a matched filter is indeed calculated through a convolution \cite{stankovic2015digital} with a time reversed feature,
\begin{gather}
	y(n)=x(n)*w(-n)=\sum_{m}x(m)w(m-n) \nonumber \\ =\sum_mx(n+m)w(m)=x(n)*_cw(n), \label{convSTand}
	\end{gather} 
where $w(n)$ denotes the feature that we are looking for in the analyzed signal, $x(n)$, and $*_c$ denotes the convolution with the time-reversed feature/template,  $w(-n)$, which serves as an  ``impulse response''. Therefore, the best search function to detect a feature, $w(n)$, within a signal $x(n)$, would be through a convolution of the signal, $x(n)$, with the time reversed feature $w(-n)$. 

\begin{Remark}
Notice that in the definition of the matched filter, physically the convolution $x(n)*w(-n)$ represents the cross-correlation of $x(n)$ and $w(n)$ rather than their convolution, $x(n)*w(n)$. This is because we have used a digital filter to implement the actual cross-correlation, which has been achieved by the impulse response within the convolution operation being a time-reversed version of the feature of interest. Despite this theoretical grounding, the open literature refers the NNs equipped with this operation as \textit{convolutional neural network} (CNN) rather than \textit{correlation neural network}. Indeed, all notations assume that the convolution is applied after one of the signals is time reversed, that is $x(n)*w(-n)$,  as in relation (\ref{convSTand}) which gives the matched filter. This is implicitly indicated in various notations in literature, for example,  $\mathbf{x}*rot 180^0 (\mathbf{w})$ or $\textrm{conv}(\mathbf{x},reverse(\mathbf{w}))$. We will use a simplified notation $\mathbf{x}*_c\mathbf{w}$, to indicate that the first signal in the convolution sum is reversed.
\end{Remark}

\begin{Remark}\label{remaMFCCNN}
Convolution based feature detection is independent of the feature position within the considered signal, since 
$y(n)=x(n)*w(-n)$
is 
calculated by sliding the window (filter/kernel), $w(n)$, which is multiplied with the actual signal segments, $x(n)$, for all $n$. 
\end{Remark}

\begin{Remark} Consider the problem of determining whether a received waveform matches any predefined waveform from a set of  waveforms of interest (dictionary). The task is to determine the best match between the received waveform and existing template waveforms. Then, it intuitively makes sense to compute the cross-correlation of the received waveform against each member of the alphabet whereby for the same normalized energy, the maximum cross-correlation corresponds to the best matching template waveform from the dictionary. One pragmatic way of calculating this set of cross-correlations is by passing the received waveform through a bank of filters,  each having as the impulse response one of the time-reversed alphabet waveforms. Then, the maximum output corresponds to the filter containing the corresponding feature signal as its impulse response. 
\end{Remark}

Such best match between the input signal and one of the existing $K$ features in the  bank of $K$ matched filters (dictionary), indicates that the $k$-th feature of interest is found as corresponding output 
\begin{equation}y_k(n)=x(n)*w_k(-n)=x(n)*_cw_k(n) \text{ for }  k=1,2,\dots,K, \label{maxdecis0}
\end{equation}
where the symbol $*_c$ denotes the convolution with a time reversed second signal in the operation (i.e., cross-correlation) $w(-n)$. 

Finally, the decision on whether the feature $k$ is contained in the input signal is based on a simple threshold operation 
\begin{equation}k=\textrm{arg}\{\max_l \{\{x(n)*_cw_l(n), \ l=1,2,\dots,K \}\}.\label{maxdecis}\end{equation}  

\begin{Remark}Since the decision in  (\ref{maxdecis}) is based on the maximum value among the outputs of the convolution filters, it will not be compromised if the negative values in (\ref{maxdecis0}) are neglected. Therefore, the following form 
$$k=\textrm{arg}\{\max_k \{\textrm{ReLU}\Big(\{x(n)*_cw_k(n)\Big), \ k=1,2,\dots,K \}\},$$
yields the same result. This form is based on the \textit{Rectified Linear Unit} (ReLU), a common  nonlinear activation function in CNNs, defined by 
$$\textrm{ReLU}(x) = \max\{0,x\}.$$
\end{Remark}

\begin{Remark}	
Notice that since we are looking for a  maximum calculated  over all $n$ samples, this operation will not be compromised if  the whole matched filter output domain is split into sub-domains  of $P$ instants; the possible local features are then found within these sub-domains. This operation underpins the so called \textit{max-pooling} operation in CNNs. 
\end{Remark}

	 Convolutional neural networks employ convolution layers, which consist of a set of convolutional filters. Different convolutional filters are typically applied over different layers, each aiming to identify a different feature in a signal.  In this way, by learning different aspects of the feature space, convolutional networks allow for a robust efficient analysis and classification of signals and images.

More complex structures of features can be taken into account by a network by employing more convolutional layers. 

\subsection{Graph convolution} 

Our perspective of convolutions being employed to implement cross-correlations in classical signal analysis shall now be directly extended to graph signals.

\textbf{System on a graph.} The input-output relation of a system with a finite impulse response (FIR) in the classical discrete-time domain is given by
$$y(n)=h_0x(n)+h_1x(n-1)+\dots+h_{M-1}x(n-M+1).$$
In a direct analogy, a system on a graph is defined using a graph shift operator,  $\mathbf{S}$, in the form \cite{stankovic2019understanding,stankovic2020data2}
\begin{equation}
	\mathbf{y}=h_0\mathbf{x}+h_1\mathbf{Sx}+\dots+h_{M-1}\mathbf{S}^{M-1}\mathbf{x}.
	\label{FIRGA}
\end{equation}
For undirected graphs, the graph Laplacian, $\mathbf{L}$, is commonly used as the graph shift operator for systems on a graph. Other graph shift operators may be used, such as the adjacency matrix, $\mathbf{A}$, and the normalized versions of the adjacency matrix, ($\mathbf{A}/\lambda_{\max}$), and the graph Laplacian ($\mathbf{D}^{-1/2}\mathbf{L}\mathbf{D}^{-1/2}$). The random walk (diffusion) matrix ($\mathbf{D}^{-1}\mathbf{W}$) is one more possible graph shift operator.

The spectral domain description of a system on a graph is obtained when the graph shift operator, for example, the graph Laplacian  matrix, $\mathbf{L}$, is presented in its eigendecomposition form 
\begin{equation}
	\mathbf{S}=\mathbf{U}\boldsymbol{\Lambda}\mathbf{U}^{-1} \text{ \ \ and \ \ } \boldsymbol{\Lambda}=\mathbf{U}^{-1}\mathbf{S}\mathbf{U},
	\label{FIRGAEIG}
\end{equation}
where $\mathbf{U}$ is the transformation matrix with the eigenvectors as its columns and $\boldsymbol{\Lambda}$ is a diagonal matrix with the corresponding eigenvalues on the main diagonal (the graph Laplacian is always diagonizable, being a real-valued symmetric matrix). By left-multiplying the vertex domain relation in (\ref{FIRGA}) by the inverse transformation matrix $\mathbf{U}^{-1}$ we obtain
\begin{gather*}
	\mathbf{U}^{-1}\mathbf{y}=h_0\mathbf{U}^{-1}\mathbf{x}+h_1\mathbf{U}^{-1}\mathbf{S}\mathbf{x}+\dots+h_{M-1}\mathbf{U}^{-1}\mathbf{S}^{M-1}\mathbf{x}.
	\label{FIRGAS}
\end{gather*}
Now, by using $\mathbf{U}^{-1}\mathbf{S}\mathbf{x}=\mathbf{U}^{-1}\mathbf{S}\mathbf{U}\mathbf{U}^{-1}\mathbf{x}=\boldsymbol{\Lambda}\mathbf{X}$, we arrive at the spectral domain description of a system on a graph \cite{stankovic2020data2}, given by
\begin{equation}
	\mathbf{Y}=h_0\mathbf{X}+h_1\boldsymbol{\Lambda}\mathbf{X}+\dots+h_{M-1}\boldsymbol{\Lambda}^{M-1}\mathbf{X},
	\label{FIRGASSS}
\end{equation}
where $$\mathbf{X}=\mathbf{U}^{-1}\mathbf{x} \text{\ \  and \ \  } \mathbf{Y}=\mathbf{U}^{-1}\mathbf{y}$$
 are the graph Fourier transforms (GFT) of the graph signals $\mathbf{x}$ and $\mathbf{y}$. The transfer function of the system, 
 $$\mathbf{Y}=H(\boldsymbol{\Lambda})\mathbf{X},$$ is a diagonal matrix defined by 
\begin{equation}
	H(\boldsymbol{\Lambda})=h_0\mathbf{X}+h_1\boldsymbol{\Lambda}\mathbf{X}+\dots+h_{M-1}\boldsymbol{\Lambda}^{M-1}. 
	\label{FIRGASSSH}
\end{equation}

\textbf{Filtering and convolutions of graph signals.} The three approaches to filtering (convolutions) of a graph signal using a system whose transfer function is $G(\boldsymbol{\Lambda})$, with the elements on the diagonal $G(\lambda_k)$, $k=1,2,\dots,N$ are as follows. 
\begin{enumerate}[label=(\alph*)]
	\item The simplest approach is based on the direct use of the GFT, and is performed by:
	\begin{enumerate}[label=(\roman*)]
		\item Calculating the GFT of the input signal, $\mathbf{X}=\mathbf{U}^{-1}\mathbf{x}$,
		\item Producing the output GFT by multiplying $\mathbf{X}$ by $G(\boldsymbol{\Lambda})$, to yield $\mathbf{Y}=G(\boldsymbol{\Lambda})\mathbf{X}$,
		\item Calculating the output (filtered) signal as the inverse DFT of $\mathbf{Y}$, that is $\mathbf{y}=\mathbf{U}\mathbf{Y}$.
	\end{enumerate}
	
	The result of this operation,  
	\begin{gather*}
		y(n)=x(n)*g(n) =\mathrm{IGFT}\{\mathrm{GFT}\{x(n)\}\mathrm{GFT}\{g(n)\}\} \\ =\mathrm{IGFT}\{X(k)G(\lambda_k)\},
		\end{gather*}
	is called \textit{a convolution of signals on a graph}.
	
	However,  this procedure quickly becomes \textit{computationally prohibitive for very large graphs} (with extremely large number of vertices $N$), since its requires $\mathcal{O}(N^2)$ operations over $N$-dimensional vectors and matrices.
	
	\item A way to avoid the full size transformation matrices for large graphs is to approximate the filter transfer function, $G(\lambda)$, at the positions of the eigenvalues,  $\lambda=\lambda_k$, $k=1,2,\dots,N$, by a polynomial, $h_0+h_1\lambda+h_{2}\lambda^{2}+\dots+h_{M-1}\lambda^{M-1}$, that is
	\begin{equation}
		h_0+h_1\lambda_k+\dots+h_{M-1}\lambda_k^{M-1}=G(\lambda_k),  \ \ \ k=1,2,\dots,N. \label{SysVV}
	\end{equation}
	The resulting system of $N$ equations 
	\begin{equation}
		\mathbf{V}\mathbf{h}=\mathrm{diag}\{G(\boldsymbol{\Lambda})\}, \label{SysVVM}
	\end{equation}
	is solved in the least squares sense for $M<N$ unknown parameters of the system, $\mathbf{h}=[h_0,h_1,\dots,h_{M-1}]^T$, with a given $M$ and $$\mathrm{diag}\{G(\boldsymbol{\Lambda})\}=[G(\lambda_1),G(\lambda_2),\dots,G(\lambda_N)]^T$$ as the column vector of diagonal elements of $G(\boldsymbol{\Lambda})$. The elements of the matrix $\mathbf{V}$ are $V(k,m)=\lambda_k^m$, $m=0,1,\dots,M-1$, $k=1,2,\dots,N$  (Vandermonde matrix).
	
	This system can efficiently be solved for a relatively small $M$ \cite{stankovic2020data2}. Then, the implementation of the graph filter is performed in the vertex domain using  $h_0,h_1,\dots,h_{M-1}$ obtained in (\ref{FIRGA}), with $\mathbf{S}=\mathbf{L}$ and the $(M-1)$-neighborhood for every considered vertex. Notice that the relation between the IGFT of $\mathrm{diag}\{G(\boldsymbol{\Lambda})\}$ and the system coefficients $h_0,h_1,\dots,h_{M-1}$ is a direct one in the classical DFT case only, while it is more complex in the general graph case  \cite{stankovic2020data2}.
	
	For large $M$, the solution to the system of equations in (\ref{SysVV}), for the unknown parameters $h_0,h_1,\dots,h_{M-1}$, can be \textit{numerically unstable due to large values of the powers} $\lambda_k^{M-1}$ for large $M$. 
	
	\item 
	Another way for avoiding the direct GFT calculation in the implementation of graph filters is by approximating the given transfer function, $G(\lambda)$, by a polynomial $H(\lambda)$ of a continuous variable $\lambda$ \cite{hammond2019spectral,behjat2019spectral,stankovic2020vertex}.
	
	\textit{This approximation does not guarantee that the transfer function $G(\lambda)$ and its polynomial approximation 
		$$H(\lambda)=h_0+h_1\lambda+\dots+h_{M-1}\lambda^{M-1}$$ will be close at a discrete set of points $\lambda=\lambda_p$, $p=1,2,\dots,N$}. However, the maximum absolute deviation of this polynomial approximation can be kept small using the so called \textit{min-max polynomials} (for example, a Chebyshev polynomial approximation of the transfer function $G(\lambda))$. After such a polynomial approximation, $H(\lambda)$, the output of the graph system, $\mathbf{Y}=H(\boldsymbol{\Lambda})\mathbf{X}$, is calculated in the vertex domain using
	$$\mathbf{y}=\Big( \sum_{m=0}^{M-1}h_m \mathbf{L}^m\Big) \, \mathbf{x}=H(\mathbf{L})\, \mathbf{x}.$$ 
	In this way, calculation of the output signal, $y(n)$, at a vertex, $n$, is localized to the input signal sample at the same vertex, $n$, and its small $(M-1)$-neighborhood. There is no need for any operation over the whole (possibly very large) graph, as in the GFT approach.      
\end{enumerate}

\begin{Remark} Some first-order systems on a graph, which are most commonly used in the GCNN, are as follows. 
	\begin{enumerate} 
		\item For the graph Laplacian as a graph shift operator, the first order system (\ref{FIRGA}) assumes the form
		\begin{gather}
			\mathbf{y}= h_0 \mathbf{x}+h_1 \mathbf{L} \mathbf{x}.
			\label{eq:filter-timeCNNGL}
		\end{gather}
	The output signal calculation requires only the samples from the one-neighborhood of every considered vertex. 
	\item The normalized graph Laplacian, 
\begin{gather*}\mathbf{L}_N=\mathbf{D}^{-1/2}\mathbf{L}\mathbf{D}^{-1/2} 
	=\mathbf{D}^{-1/2}(\mathbf{D}-\mathbf{W})\mathbf{D}^{-1/2} \\
	=\mathbf{I}-\mathbf{D}^{-1/2}\mathbf{W}\mathbf{D}^{-1/2}=\mathbf{I}-\mathbf{W}_N,
	\end{gather*}
	where $\mathbf{W}_N=\mathbf{D}^{-1/2}\mathbf{W}\mathbf{D}^{-1/2}$, is commonly used as a shift operator in the first-order system to define the convolution and the convolution layer in the graph convolutional neural networks (GCCN) \cite{gori2005new, scarselli2008graph}. Its form is
	\begin{gather}
		\mathbf{y}=( h_0 \mathbf{L}_N^0+h_1 \mathbf{L}_N^1)  \, \mathbf{x}  =(h_0+h_1) \mathbf{x}-h_1\mathbf{W}_N\mathbf{x}.
		\label{eq:filter-timeCNN}
	\end{gather}  
	\item \textbf{Multichannel systems.} Relation (\ref{eq:filter-timeCNN}) can be adapted to  include  the channel index, $k$, in the case of multichannel systems. The  $k$-th channel in the GCNN convolutional layer, with the input signal, $\mathbf{x}$, and the output, $\mathbf{y}_k$, is implemented as
	\begin{gather}
		\mathbf{y}_k=w_{k}(0) \mathbf{x}+w_{k}(1)\mathbf{D}^{-1/2}\mathbf{W}\mathbf{D}^{-1/2}\mathbf{x} \nonumber \\
		=w_{k}(0) \mathbf{x}+w_{k}(1)\mathbf{W}_N\mathbf{x},
		\label{eq:filter-timeCNN2222}
	\end{gather}  
	for $k=1,2,\dots,K$, where the weights $w_{k}(0)$  and $w_{k}(1)$ in the $k$-th channel, correspond respectively to the weights $(h_0+h_1)$ and $(-h_1)$ in  (\ref{eq:filter-timeCNN}). Further simplification of (\ref{eq:filter-timeCNN2222}), by using just one parameter, $w_{k}(0)=w_{k}(1)=\theta_k$, was originally proposed for the GCNN  \cite{gori2005new, scarselli2008graph}. Since this may over-reduce the parameter space in GCNNs, we will resort to two parameters, $w_{k}(0)$  and $w_{k}(1)$. 
	
	\item The \textit{random walk operator} produces the first-order multichannel system of the form \begin{gather}
		\mathbf{y}_k= w_k(0) \mathbf{x}+w_k(1) \mathbf{D}^{-1}\mathbf{W} \mathbf{x},
		\label{eq:filter-timeCNNGL}
	\end{gather}
where $k=1,2,\dots,K$. However, this graph shift operator does not preserve the symmetry property of the shift matrix.

 \item For \textit{directed} unweighted graphs, graph shift in the form of the adjacency matrix can be used. The system on a graph signal for one channel is then given by 
 \begin{gather}
 	\mathbf{y}_k= w_k(0) \mathbf{x}+w_k(1) \mathbf{A} \mathbf{x}+w_k(2) \mathbf{A}^T \mathbf{x},
 	\label{eq:filter-timeCNNGLAA}
 \end{gather}
where $\mathbf{A}\mathbf{x}$ denotes the graph backward shift, and $\mathbf{A}^T\mathbf{x}$ is used for the forward shift on a graph. 

Note that we can equally use the normalized adjacency matrix, $\mathbf{A}/\lambda_{\max}$, instead of the adjacency matrix, $\mathbf{A}$.
	\end{enumerate}
\end{Remark}

\subsection{Graph Matched Filter}

To derive the matched filter form within the graph signal framework, consider the output of a system, defined by the graph transfer function, $G(\lambda_k)$, and the input graph signal, $x(n)$, that is 
\begin{gather*}y(n)=x(n)*g(n)=\mathrm{IGFT}\{X(k)G(\lambda_k)\}\\ =\sum_{k=1}^NX(k)G(\lambda_k)u_k(n).
	\end{gather*} 
In matched filtering, for a given signal form, $x(n)$, the aim is to find $g(n)$ or $G(\lambda_k)$ that maximizes the output signal value, $y(n)$, at a vertex $n=n_0$. The squared absolute value of $y(n_0)$ is then defined by
\begin{gather*}|y(n_0)|^2=\Big|\sum_{k=1}^NX(k)G(\lambda_k)u_k(n_0)\Big|^2 \\ 
	\le \sum_{k=1}^N|X(k)|^2 \sum_{k=1}^N|G(\lambda_k)u_k(n_0)|^2,
	\end{gather*} 
according to the Schwartz inequality; the maximum is achieved when the equality holds. For real-valued functions, this is true when the condition
\begin{equation}G(\lambda_k)u_k(n_0)=X(k) \label{GMFil}
	\end{equation}
holds up to a possible multiplication (scaling) constant. The maximum absolute squared value of the output,  given by
$$|y(n_0)|^2=\Big(\sum_{k=1}^N|X(k)|^2\Big)^2=E_x^2,$$
is therefore achieved when (\ref{GMFil}) holds, where $E_x=|y(n_0)|$ is the input signal energy. 

\begin{Remark}
In classical analysis (when the corresponding GFT is complex-valued \cite{stankovic2020data1}), it is well known that
$$G(\lambda_k)=X^*(k)/u_k(n_0)=X^*(k)e^{-j2\pi n_0k/N}/\sqrt{N}$$
 or for the matched filter impulse response with $n_0=0$ \cite{stankovic2015digital} 
 $$g(n)=x^*(-n).$$ However, in the case of general graphs and graph signals, the vertex domain form of (\ref{GMFil}) is much more complicated. 
\end{Remark}

 Next, we proceed to analyze more general graph forms starting with a special case when the input signal can be considered as a result of a \textit{diffusion process}.

\textbf{Graph matched filter for a diffusion signal.} The intuition and implementation of the graph matched filter can be significantly simplified if we assume that the considered signal is a result of an $(M-1)$-step diffusion process, with the initial signal to the diffusion being a unit delta pulse at an arbitrary vertex $n_0$.

   Consider a graph signal that is obtained by an  $(M-1)$ step diffusion from a delta pulse signal, $\mathbf{x}_0$, at a vertex $n_0$, whose element-wise form is $x_0(n)=\delta(n-n_0)$. Then, the resulting graph signal, $\mathbf{x}$, is given by 
\begin{equation}
	\mathbf{x}=a_0\mathbf{x}_0+a_1\mathbf{L}_N\mathbf{x}_0+\dots+a_{M-1}\mathbf{L}^{M-1}_N\mathbf{x}_0. \label{GFTgrFet}
\end{equation}  
where $a_{\nu}$, $\nu=0,1,\dots,M-1$, are the diffusion constants, 
while the GFT of $\mathbf{x}_0$ is defined as
$$X_0(k)=\sum_{n=1}^Nx_0(n)u_k(n)=u_k(n_0).$$
The GFT of the graph signal, $\mathbf{x}$, follows from (\ref{GFTgrFet}) and (\ref{FIRGASSS}) as 
\begin{gather}\mathbf{X}=\Big(a_0+a_1\mathbf{\boldsymbol{\Lambda}}_N+\dots+a_{M-1}\mathbf{\boldsymbol{\Lambda}}^{M-1}_N\Big)\mathbf{X}_0
	\end{gather} 
or in an element-wise form
\begin{gather}
	X(k)=\Big(a_0+a_1\lambda_k+\dots+a_{M-1}\lambda^{M-1}_k\Big)u_k(n_0).
\end{gather} 

For this class of graph signals, the spectral domain solution of (\ref{GMFil}) follows straightforwardly in the form
 \begin{gather}
 	G(\lambda_k)=a_0+a_1\lambda_k+\dots+a_{M-1}\lambda^{M-1}_k.
 \end{gather} 
Although the vertex domain form of $g(n)$ in $y(n)=x(n)*g(n)$ may be quite complex, the vertex domain implementation of this matched filter is rather  simple and  follows from $Y(k)=G(\lambda_k)X(k)$ to yield
\begin{gather}
\mathbf{y}=\Big(a_0+a_1\mathbf{L}_N+\dots+a_{M-1}\mathbf{L}_N^{M-1}\Big) \mathbf{x}.
\end{gather} 

The relation between the matched filter output and the initial delta pulse signal in the diffusion system is obtained from the above for $\mathbf{x}=\mathbf{x}_0$.
 
\begin{Remark}
The vertex domain form of the graph matched filter is the inverse GFT of the graph filter transfer function, $G(\lambda_k)$, given by
$$g(n)=\sum_{k=1}^NG(\lambda_k)u_k(n).$$
Denote by 
$$g_0(n)=\sum_{k=1}^Nu_k(n)$$
and by $\mathbf{g}_0$ a vector whose elements represent sums of all eigenvector elements, $u_k(n)$, at a given vertex $n$, or in matrix form 
$$\mathbf{g}_0=\mathbf{U}\mathbf{1}$$
where $\mathbf{1}$ is a column vector with all elements equal to 1.  In general, the elements $g_0(n)$ are nonzero for all $n$.

The matrix form of the vertex domain matched filters then becomes 
\begin{gather*}\mathbf{g}=\mathbf{U}G(\mathbf{\boldsymbol{\Lambda}})\mathbf{1}=\mathbf{U}\Big(a_0+a_1\mathbf{\boldsymbol{\Lambda}}_N+\dots+a_{M-1}\mathbf{\boldsymbol{\Lambda}}^{M-1}_N\Big)\mathbf{1}\\
	=a_0\mathbf{g}_0+a_1\mathbf{L}_N\mathbf{g}_0+\dots+a_{M-1}\mathbf{L}_N^{M-1}\mathbf{g}_0.
	\end{gather*}
\end{Remark}
In  classical analysis, with the adjacency matrix on a directed circular graph serving as a shift operator \cite{stankovic2020data2}, we immediately arrive at the impulse response of the classical matched filter, given by 
\begin{gather*} 
	g_0(n)=\sum_{k=1}^Nu_k(n)=\sum_{k=1}^Ne^{j2\pi nk/N}/\sqrt{N}=\delta(n)\sqrt{N},  \\ 
	\lambda_k=e^{-j2\pi k/N}, \  \  \text{ and } \\ g(n)=a_0\delta(n)+a_1\delta(n-1)+\dots+a_{M-1}\delta(n-M+1).
	\end{gather*}

\begin{figure*}
	\centering
	
	\includegraphics[scale=0.9]{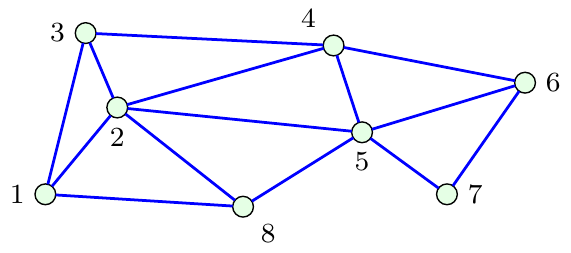}	\hspace{10mm}
	\includegraphics[scale=0.9]{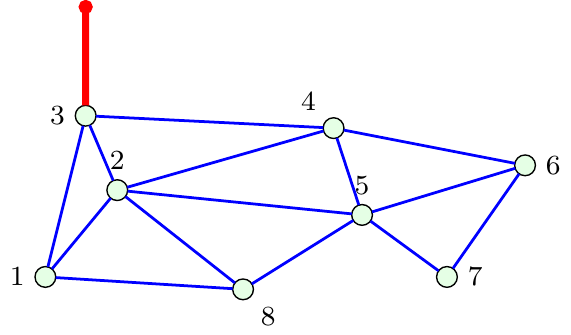}
	\\
	(a) \hspace{65mm} (b)
	\\ \vspace{3mm} 
	\includegraphics[scale=0.9]{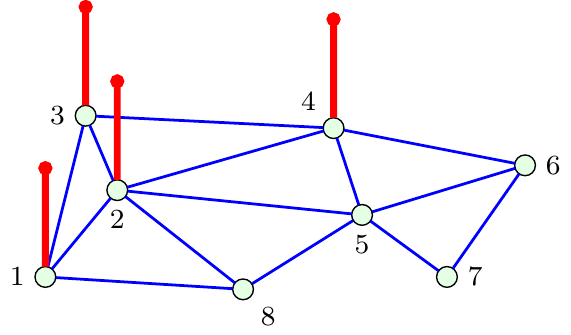}
	\hspace{10mm}
	\includegraphics[scale=0.9]{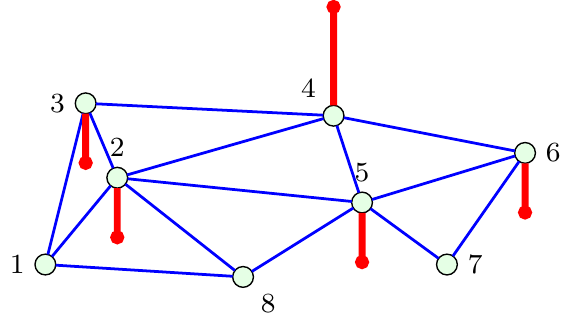}
	\\
	(c) \hspace{65mm} (d)
	\\ \vspace{3mm} 
	\includegraphics[scale=0.9]{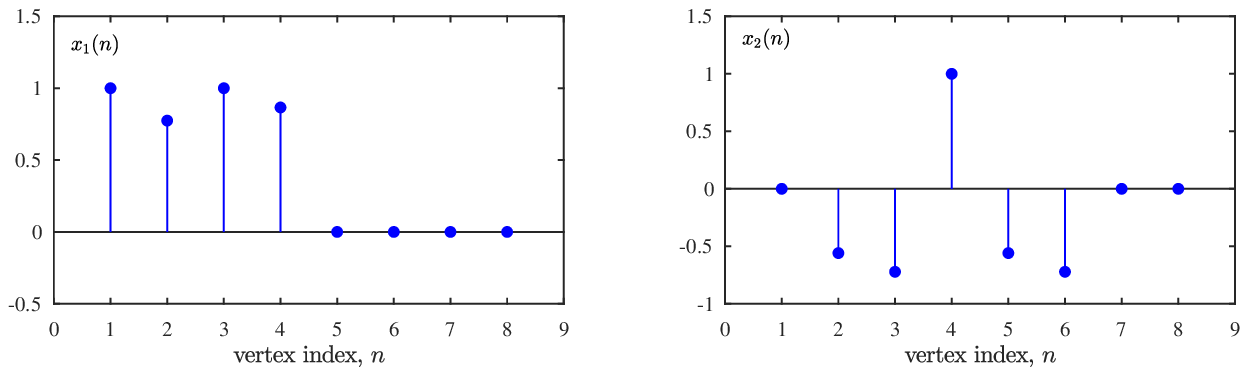} \\
	(e) \hspace{65mm} (f)
	\\ \vspace{3mm} 
	\includegraphics[scale=0.9]{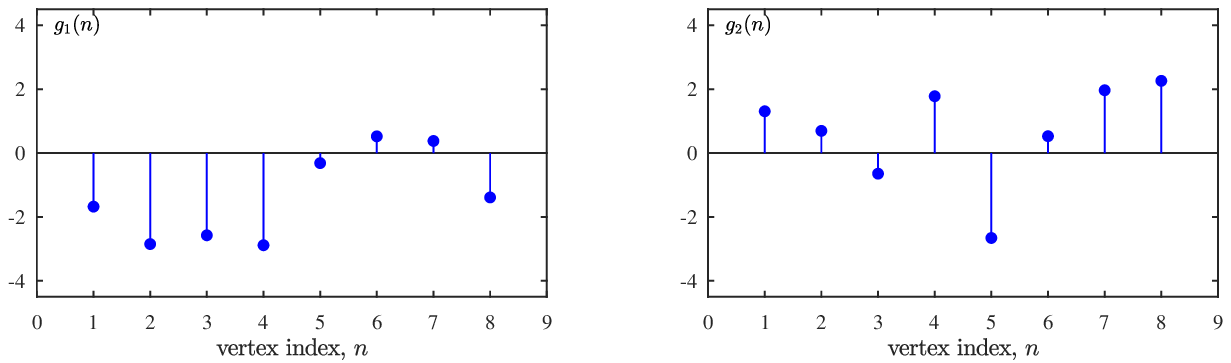} \\
	(g) \hspace{65mm} (h)
	\\ \vspace{3mm} 
	\includegraphics[scale=0.9]{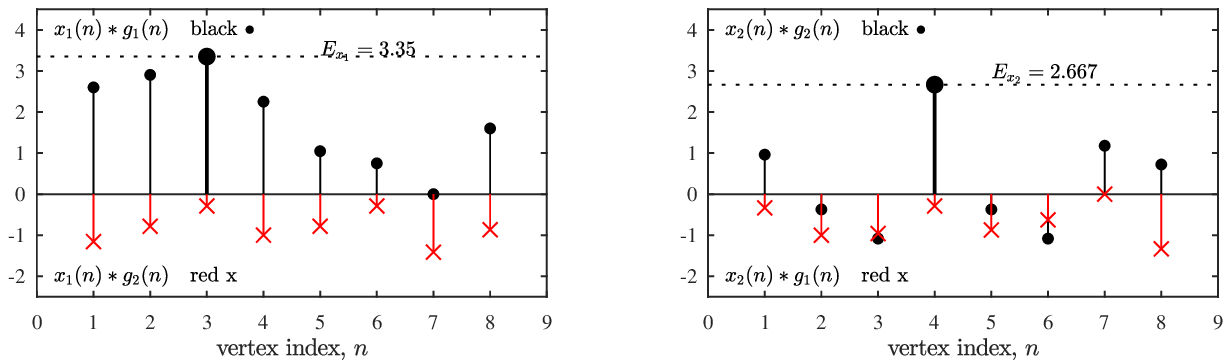}	 \\
	(i) \hspace{65mm} (j)
	\caption{Two signals (features), $\mathbf{x}_1$ and $\mathbf{x}_2$ observed on an undirected unweighted graph. (a) Considered graph topology.  (b) Unit pulse graph signal, $x_0(n)=\delta(n-3)$, at vertex $n=3$. (c) The first feature is obtained by shifting the unit pulse $x_0(n)=\delta(n-3)$ from vertex $n=3$ to its one-neighborhood with the weight of $3$, that is, $\mathbf{x}_1= \mathbf{x}_0+3\mathbf{W}_N\mathbf{x}_0$. (d) The second feature is obtained by shifting the unite pulse, $x_0(n)=\delta(n-4)$, to its one-neighborhood vertices, with the shift weight of $-2$, that is $\mathbf{x}_2= \mathbf{x}_0-2.5\mathbf{W}_N\mathbf{x}_0$. The graph signals from panels (c) and (d) are respectively given on a liner vertex-index axis in (e) and (f). 
	The matched filter responses $\mathbf{g}_1=4\mathbf{g}_0-3\mathbf{L}_N\mathbf{g}_0$  and
	$\mathbf{g}_2=-1.5\mathbf{g}_0+2.5\mathbf{L}_N\mathbf{g}_0$ are shown in panels (g) and (h).	
		The result of the graph matched filter with the adjusted impulse response $g_1(n)$ corresponding to $x_1(n)$ and $g_2(n)$ corresponding to $x_2(n)$, are given in panels (i) and (j), respectively, denoted by black dots. The cross-correlations of  $x_1(n)$ and  $g_2(n)$ as well as $x_2(n)$ and  $g_1(n)$ are designated with red crosses in both panels.}
	\label{fig-sig-gr1}
\end{figure*}

\begin{figure}[tbp]
	\centering
	\includegraphics[]{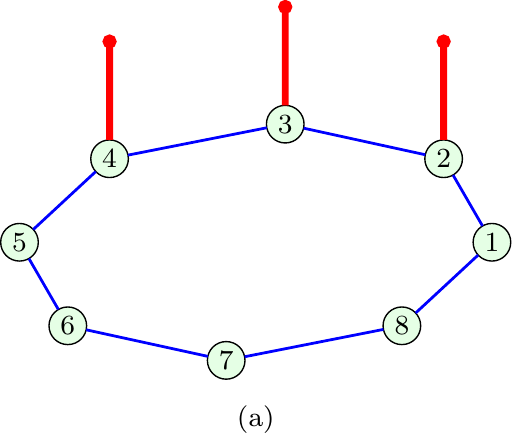}
	\hspace{10mm}
	\includegraphics[]{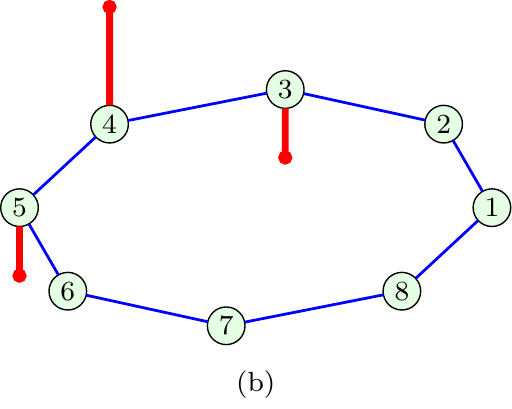}
	\caption{The two signal features, $\mathbf{x}_1=[0,1,1,1,0,0,0,0]^T$ and $\mathbf{x}_2=[0,0,-0.5,1,-0.5,0,0,0]^T$ shown on an undirected unweighted circular graph; notice that this corresponds to the classical “regularly sampled” time-domain analysis. The first feature is obtained by shifting the unit pulse, $x(n)=\delta(n-3)$, to the left and right with the shift weight $2$, that is $\mathbf{x}_1= \mathbf{x}+2\mathbf{D}^{-1/2}\mathbf{W}\mathbf{D}^{-1/2}\mathbf{x}$. The second feature is a result of  the unit pulse, $x(n)=\delta(n-4)$, being shifted with the shift weight $-1$ using the normalized adjacency (weight) matrix, that is $\mathbf{x}_2= \mathbf{x}-\mathbf{D}^{-1/2}\mathbf{W}\mathbf{D}^{-1/2}\mathbf{x}=\mathbf{x}-\mathbf{W}_N\mathbf{x}$. }
	\label{fig:sig-arb-graph}
\end{figure}

\textit{Calculation of the graph matched filter responses} is much more involved and will be illustrated on a simple undirected graph with unit edge weights, shown in Fig. \ref{fig-sig-gr1}(a). This graph is used as an irregular domain signal domain for the considered analysis (the weight matrix, $\mathbf{W}$, of this graph is given  by (\ref{matA1a})).
\begin{exmpl}
	 Consider two graph signals (with corresponding features) on the graph shown in Fig. \ref{fig-sig-gr1}(a) that are obtained by graph shifting a delta pulse, $x_0(n)=\delta(n-n_0)$, from a vertex $n_0$ to its neighborhood in two ways as below.
\begin{enumerate}[label=(\roman*)]
\item The first signal (feature) is produced by a graph shift given by
$$
	\mathbf{x}_1= \mathbf{x}_0+3\mathbf{W}_N\mathbf{x}_0 = 4 \mathbf{x}_0-3\mathbf{L}_N \, \mathbf{x}_0, 
	$$
with  $x_0(n)=\delta(n-3)$, shown in Fig. \ref{fig-sig-gr1}(b);
\item The second signal (feature) results from
$$
	\mathbf{x}_2=\mathbf{x}_0-2.5\mathbf{W}_N\mathbf{x}_0 = -1.5 \mathbf{x}_0+2.5\mathbf{L}_N \, \mathbf{x}_0,  
	\label{eq:filter-timeCNN2ex}
$$
with $x_0(n)=\delta(n-4)$ and $\mathbf{L}_N=\mathbf{I}-\mathbf{W}_N$, where 
\begin{gather}
	\!\!\mathbf{W}_N=
	\begin{array}{cr}
		\begin{bmatrix}
			  \vspace{0.3mm} \ 0\  &  0.26 \ &  \!\!\!\! \phantom{-}\tfrac{1}{3} \  & \ 0\  & \ 0\  & \ 0\  & \ 0\  &   \!\!\! \phantom{-}\tfrac{1}{3} \  \\ \vspace{0.3mm}
			 0.26\  & \ 0\  &  0.26 \  &   0.22  &  \!\!\!\! \phantom{-}\tfrac{1}{5} \  & \ 0\  & \ 0\  &  0.26    \\ \vspace{0.3mm}
			 \!\!\! \phantom{-}\tfrac{1}{3}\  &  
			 0.26  & \ 0\  &  0.29  &  \ 0\  & \ 0\  & \ 0\  & \ 0\   \\ \vspace{0.3mm}
			\ 0\  &  0.22  & 0.29 & \ 0\  & 0.22  &  0.29  & \ 0\  & \ 0\   \\ \vspace{0.3mm}
			\ 0\  & \!\!\!\!\!  \phantom{-}\tfrac{1}{5}\  & \ 0  &  0.22  & \ 0\  & 0.26  & 0.32  & 0.26   \\ \vspace{0.3mm}
			\ 0\  & \ 0\  & \ 0\  & 0.29  & 0.26  & \ 0\  & 0.41  & \ 0\   \\ \vspace{0.3mm}
			\ 0\  & \ 0\  & \ 0\  & \ 0\  & 0.32  & 0.41  & \ 0\  & \ 0\   \\ \vspace{0.3mm}
			\!\!\! \phantom{-}\tfrac{1}{3}\  & 0.26  & \ 0\  & \ 0\  & 0.26  & \ 0\  & \ 0\  & \ 0\ 
		\end{bmatrix} 
	\end{array}\!\!. \label{matA1aL}
\end{gather}
\end{enumerate}
  
The signals $\mathbf{x}_1$ and $\mathbf{x}_2$ are shown on the corresponding graph in Fig. \ref{fig-sig-gr1}(c) and (d), while a vertex-index axis representation is given in Fig. \ref{fig-sig-gr1}(e) and (f).

The graph impulse responses of the corresponding matched filters are given in Fig.  \ref{fig-sig-gr1}(g) and (h) and assume the form
\begin{gather}
	\mathbf{g}_1=4\mathbf{g}_0-3\mathbf{L}_N\mathbf{g}_0,   \\
	\mathbf{g}_2=-1.5\mathbf{g}_0+2.5\mathbf{L}_N\mathbf{g}_0	
	\end{gather}
where $g_0(n)=\sum_{k=1}^Nu_k(n)$. 

 The system responses (outputs) to the input signals, $\mathbf{x}_1$ and $\mathbf{x}_2$, are calculated using  the vertex domain forms of both filters, $\mathbf{g}_1$ and $\mathbf{g}_1$, to yield  
 \begin{gather}
 	\mathbf{y}_i=4\mathbf{x}_j-3\mathbf{L}_N\mathbf{x}_j \\
 	\mathbf{y}_i=-1.5\mathbf{x}_j
 	+2.5\mathbf{L}_N\mathbf{x}_j,
 	\end{gather}
 for $i,j=1,2$. The results were next verified by the corresponding spectral domain relations \begin{gather}y_i(n)=x_i(n)*g_j(n)=\mathrm{IGFT}\{GFT\{x_i(n)\}GFT\{g_j(n)\}\}\}. \nonumber
 	\end{gather}
  While these two forms produce the same results, the vertex domain relation is simpler for realization since it  only uses the local signal samples from one-neighborhoods.
 
The corresponding  matched filter responses are given in Fig. \ref{fig-sig-gr1}(i) and (j). The maximum values of the matched filter responses are checked against the corresponding graph signal energy.  

For direct comparison with the classical time domain matched filters and features, similar signals are presented on an undirected and unweighted circular graph in Fig. \ref{fig:sig-arb-graph}.

\end{exmpl}

\section{Forward Propagation}\label{forwardP} 

We shall now use a matched filter perspective to shed a new light on key algorithmic steps in the operation of GCNNs. For simplicity we assume that the weights of the convolutional filters (\textit{forward propagation}) are  already initialized or calculated in some other way. The weight update will be addressed afterwards.

\begin{enumerate}
	
	\item \textbf{Input:}  Consider a graph signal, $\mathbf{x}$, the samples of which are observed at $N$  vertices, and given by
	$$\mathbf{x}=[x(1), \ x(2), \ \dots, x(N)]^T.$$
	
	A common goal in GCNNs is to classify input signals into several non-overlapping categories of signals containing specific features.
	
	\item \textbf{Convolution layer:} This operation employs a convolutional filter of $M$ elements. The convolution layer corresponds to a graph matched filter and is sometimes called the \textit{convolutional kernel}.  Note that $K$ such graph matched filters are applied, if we are looking for $K$ features in $\mathbf{x}$. The  elements of the $k$-th kernel of the graph matched filter (graph convolutional layer) are then 
	$$\mathbf{w}_k=[w_k(0), \ w_k(1), \ \dotsc, \ w_k(M-1)]^T.$$
	for $k=1,2,\dots,K$.
	 A commonly used  choice for GCNN is $M=2$. Using the normalized weight matrix as a shift operator, the output signals from the graph matched filters, for $M=2$, are obtained from (\ref{eq:filter-timeCNN2222}) and given by
	 \begin{gather}
	 \mathbf{y}_k=\mathbf{w}_k*\mathbf{x}=w_{k}(0) \mathbf{x}+w_{k}(1)\mathbf{D}^{-1/2}\mathbf{W}\mathbf{D}^{-1/2}\mathbf{x} \nonumber \\ =w_{k}(0) \mathbf{x}+w_{k}(1)\mathbf{W}_N\mathbf{x} \label{outWn}
	\end{gather}  
	where $*$ denotes the graph convolution (cross-correlation) or the matched filter response (feature) $\mathbf{w}_k$, and the  signal, $\mathbf{x}$.  
		For $M=3$, the matched filter channels would be implemented as
	\begin{equation}
	\mathbf{y}_k=\mathbf{w}_k*\mathbf{x}=w_{k}(0) \mathbf{x}+w_{k}(1)\mathbf{W}_N\mathbf{x}+w_{k}(2)\mathbf{W}^2_N\mathbf{x},	
	\label{ConvCNN3}
	\end{equation}
	with the dimension of the $k$-th output, $\mathbf{y}_k$, being $N \times 1$.  
	
	In total, $K$ such output signals of the graph convolution layer, $\mathbf{y}_k$, $k=1,2\dots,K$, are obtained, with the total number of the output signal elements, $y_k(n)$, from the graph convolution layer therefore being $KN$.  
	
	Relation (\ref{outWn}) can also be written in terms of the normalized Laplacian as a shift operator, with the corresponding coefficients $h_k(m)$, in the form
	\begin{gather} 
		\mathbf{y}_k=\mathbf{h}_k*\mathbf{x}=h_{k}(0) \mathbf{x}+h_{k}(1)\mathbf{L}_N \mathbf{x}.
		\label{eq:filter-timeCNN222}
	\end{gather} 
	
	\begin{Remark}
	The total number of filter weights, $w_k(\nu)$, $k=1,2,\dots,K$, $\nu=0,1,\dots,M-1$, in the graph convolutional layer is equal to the product of the graph convolution filter length, $M$, and the number of convolutional (matched) filters, $K$, that is $MK$. This is typically much smaller than in the case of a fully connected neural network, whereby for every layer, each of the $N$ input signal samples is connected through weights to each of $K$ output signals, a total of $NK$ connections. 
	\end{Remark}

	\begin{exmpl}
		\textbf{Relation to standard convolutional neural networks.} The input-output relation for the GCNN simplifies into a standard CNN as a special case for  circular undirected and unweighted graph. This is immediately seen by first considering the element-wise form of the output, given by \cite{CNN_Tutorial}
		\begin{gather*}
		y(n)= (w_{k}(0)+\frac{1}{2})]x(n)+ w_{k}(1)\frac{1}{2}[x(n+1)+x(n-1)] \\ 
		+  w_{k}(2)\frac{1}{4}[x(n+2)+x(n-2)], \label{ConvNNNN}
		\end{gather*}
		where the factors $1/2$ arise due to the degree matrix, and can be absorbed into the weights, $w_{k}(\nu)$. 
		Since the considered graph is undirected, the above form is symmetric. An asymmetric form of the standard CNN could be obtained by using the adjacency matrix, $\mathbf{A}$, of a directed unweighted circular graph, instead of the normalized weight matrix, $\mathbf{W}_N$.  The relation  in (\ref{eq:filter-timeCNNGLAA}) then produces the standard CNN convolution
		\begin{gather*}
			y_k(n)= w_{k}(0)x(n)+ w_{k}(1)x(n-1)
			+  w_{k}(2)x(n+1). \label{ConvNNNNA}
		\end{gather*}
		\end{exmpl}   	
	
	\item \textbf{Bias:}  A bias (constant) term may be added at the graph convolution layer (like in  standard neural network layers),  to yield
	$$\mathbf{y}_k=\mathbf{w}_k*\mathbf{x}+b_k,$$
	whereby the total number of coefficients in every convolution is increased by one. 

	\item \textbf{Nonlinear activation function:} Signals and images are far from exhibiting a linear nature, while convolution (correlation) is a linear operation. To this end, a non-linearity is applied to the output of a convolutional layer.  The most common  nonlinear activation function for CNNs and GCNNs is the Rectified Linear Unit (ReLU), defined by 
\begin{equation}f(y) = \max\{0,y\}.\label{ReLUdef}\end{equation}
	In GCNNs, this function has several advantages over sigmoidal-type activation functions: (i) it does not saturate for  positive values of input, thus producing nonzero gradient for large input values, (ii) its calculation is not computationally demanding, and (iii) in practical applications ReLU converges faster than  saturation-type nonlinearities (logistic, tanh).  Moreover, this function does not activate all neurons at the same time, so that “sparsification by deactivation” is achieved for each neuron producing a negative value as an input to the ReLU activation function.       
	
	The output of the graph convolutional layer, after the activation function, then becomes
	$$f(\mathbf{y}_k)=f(\mathbf{w}_k*\mathbf{x}+{b}_k)=f(\mathbf{w}_k,\mathbf{x}).$$

%
%
%
%
%
	
	Since the ReLU is defined in such a way that it produces a zero output for negative input values, the main problem with the ReLU activation function arises when the input to a  neuron has many consecutive negative values so that the corresponding zero-output of the ReLU function will leave this neuron without an update of its weights (“dying ReLU”). This problem can be avoided using the Leaky ReLU function, whereby negative values of the input are mapped onto small scaling factors, for example, $f(y_k(n))=0.01y_k(n)$, for $y_k(n)<0$.

	\item \textbf{Pooling:} In order to reduce a possibly excessive size of the data throughput, the output signals at each layer are typically further down-sampled through the so called \textit{pooling} operation, with the output signal assuming the form
		$$\mathbf{o}_k=F\Big(f(\mathbf{w}_k*\mathbf{x}+{b}_k)\Big).$$ The max-pooling in GCNN is closely related to downscaling of the considered graph and	
	reduces the  size of the representation, and thus helps decrease the  computation requirements and the number of weights in a GCNN. One approach to down-scaling a graph is termed graph coarsening, and is explained in  Appendix A. 	
	
	\item \textbf{Flattening:} One-dimensional output signals, $\mathbf{o}_k$, after the ReLU and possibly pooling operations, are already in a vector form. These vectors are then concatenated to form the vector $\mathbf{o}_F$,
	 the elements of which are given by 
	$$o_F(m)=o_F\Big((k-1)N+n\Big)=o_k(n),  $$
 for $k=1,2,\dots,K$, $n=1,2,\dots,N$, $m=1,2,\dots,KN$.
	This process is called the \textit{flattening} operation. The  vector $\mathbf{o}_F$ is of size $KN$ if no max-pooling is performed.  If max-pooling with a factor of $P$ is used, the size of the concatenated (\textit{flattened}) vector $\mathbf{o}_F$ is $KN/P.$

	\item \textbf{Repeated graph convolutions:}  Notice that before  the flattening operation, the convolution  steps can be repeated one or more times, using different sets of filter functions. Repeated convolutions help to find more complex possibly \textbf{hierarchically composed features}. The convolutional steps can be repeated with or without the activations and pooling functions, that is, the \textit{convolution-activation-pooling}.

	\item \textbf{Fully Connected (FC) Layers:} The outputs of the previous convolutional steps, after flattening, are connected in the same form of flattened data to a standard neural network layer with fully connected neurons.  The FC layers may have a traditional \textit{multilayer structure}, and are followed by the output layer.
	
	For illustration, Fig. \ref{fig-sig-gr1-CNN} shows a simple GCNN structure, with 
	\begin{itemize}
		\item
	A graph input signal, $\mathbf{x}$, on a graph with the weight matrix, $\mathbf{W}$; 
	\item 
	One convolutional layer with weights, $w_k(\nu)$, $K=2$, and $M=2$; 
	\item One fully connected standard neural network layer with $KN=16$ input neurons and $S=2$ output neurons, and
	\item Softmax output layer with $S=2$ outputs.
	\end{itemize}
This model is used in the  sequel in our numerical example to illustrate  the principle of GCNN. 
		
	\begin{figure}[hptb]
		\begin{center}
			\includegraphics{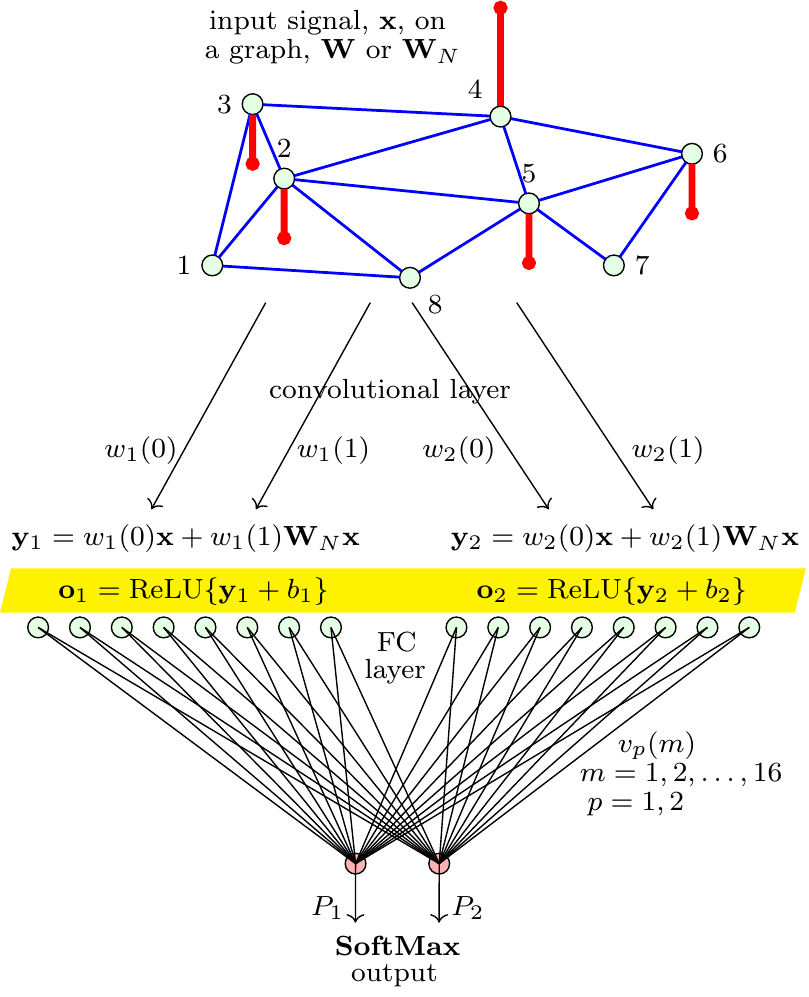} 
			\caption{Principle of a GCNN. The operation is illustrated based on a GCNN with one graph convolution layer and one FC layer, and  two neurons at the output (softmax) layer. }%
			\label{fig-sig-gr1-CNN}%
		\end{center}
	\end{figure}

\end{enumerate} 

\subsection{Updating graph convolution weights: Back-propagation}\label{BackPropag}

The initial parameters (weights) of a GCNN are typically updated in a supervised way through a gradient-based learning process known as the back-propagation (BP). For each iteration of the BP algorithm, the gradient value (sensitivity) for each network parameter (weights in the convolutional layers, weights in the fully-connected layers, and biases) is computed. These sensitivities are then used to iteratively update all the GCNN parameters until a certain stopping criterion is met or the training data set is exhausted. 

\subsubsection{Initialization}
 Unlike standard adaptive systems where the initial weight values are typically   set to zero,  the initial values of the weights in neural networks  are typically set as random (and different) values for each channel and layer.  Since the graph convolutional weights, $w_k(\nu)$, multiply, in general, $M$ input signal values in each channel (at the considered input neurons of the layer), the only requirement is that the choice of the initial weights preserve the expected energy of the output for the considered layers. This  is achieved, for example,  if the initial weights are \textit{Gaussian distributed}, with 
\begin{equation}w_k(\nu) \sim \sqrt{\frac{2}{M}}\mathcal{N}(0,1), \ \ \nu=0,1,\dots,M-1, \ \ k=1,2,\dots,K.\label{InitieighG}\end{equation}
The factor of 2 is used since the ReLU activation function will remove negative output values, which accounts for half of the expected energy. 

Another possibility is to use \textit{uniformly} distributed initial wights, $w_k(\nu)$, whereby the sum of $M$ initial weights, $\sum_{\nu=0}^{M-1} w_k(\nu)$, is also a random variable with unit variance. Such uniformly distributed random weights are defined on the interval
$$w_k(\nu) \sim \Big[-\sqrt{\frac{6}{M}}, \ \sqrt{\frac{6}{M}}\Big].$$
with the variance,  $\mathrm{Var}\{w_k(\nu)\}=\frac{6}{M}\frac{1}{3}$. In this way, the variance of a sum of $M$ values, divided by 2, to  account for the ReLU property, produces unit weight variance. The so produced initial weights  are referred to as the He initial values. 

For the fully connected layers, we can use the initial weights as for $w_k(\nu)$, with the number of input neurons of $KN$ instead of $M$. Values of the initial weights in (\ref{InitieighG}) can also be reduced, by taking into account the number of output neurons for the considered layer, $S$, to yield the Xavier  initial values, given by
$$v_p(m) \sim \sqrt{\frac{2}{NK+S}}\mathcal{N}(0,1).$$

\subsubsection{Back-propagation in a two-layer GCNN}
Consider the weight update in a simple GCNN which comprises two layers, a convolutional layer and one fully connected output layer, as shown in Fig. \ref{fig-sig-gr1-CNN}. 

\noindent\textbf{Convolutional layer.} For the input  graph signal, $\mathbf{x}=[x(1), \ x(2), \dots, x(N)]^T$, the output signal of the convolutional layer of the GCNN, with $K$ filters of the width $M=2$, is given by
\begin{align}
	\mathbf{y}_k & =w_{k}(0) \mathbf{x}+w_{k}(1)\mathbf{W}_N\mathbf{x},\nonumber
 \end{align}
	or element-wise  
	\begin{align}
y_k(n) & =w_k(0)x(n)+w_k(1)\sum_{\mu}W_N(n,{\mu}) x({\mu}),	
\label{ConvCNNK}
\end{align}
for the channels $k=1,2,\dots,K$, where $W_N(n,\mu)$ are the elements of the normalized weight matrix, $\mathbf{W}_N$.  The overall output of the convolution layer is then obtained after the bias term, $b_k$, is included and upon the application of the ReLU activation function, $f(\cdot)$, to yield
\begin{equation}o_k(n)=f\Big(y_k(n)+b_k\Big).
\label{ConvCNN)}
\end{equation}
For simplicity, we assumed that no max-pooling or any other down-sampling is performed. 

The output from the convolutional layer is then reshaped into a vector of length $KN$, which serves as input to the fully connected (FC) layer with  $S$ outputs, with only one FC layer.
Each of the $KN$ nodes of the output of the convolutional layer, $o_F(m)$, with the corresponding samples, 
$$[o_1(1),\dots,o_1(N), \ o_2(1),\dots,o_2(N), \dots, o_K(1),\dots,o_K(N)]^T$$ 
is connected  to each of the $S$ nodes, $p=1,2,\dots,S$, of the fully connected output layer, through the weights $v_p(m)$,  to produce the overall GCNN output of the form
\begin{align}z_p & =v_{p}(0)o_1(1)+v_{p}(1)o_1(2)+\dots+v_{p}(N-1)o_1(N) \nonumber  \\& +v_{p}(N)o_2(1)+ \ \ \ \  \ \ \ \ \dots \ \ \  \ \ \ \ \ +v_{p}(2N-1)o_2(N) \nonumber  \\
& \vdots  \nonumber  \\
& +v_{p}((K-1)N)o_K(1)+ \dots +v_{p}(KN-1)o_K(N).
\label{ConvCNNFC}
\end{align}
 Note that the number of weights, $v_{p}(m)$, $p=1,2,\dots,S$, $m=1,2,\dots,KN$ in the FC layer is $S\times KN$.

A commonly used \textit{loss function} in the BP algorithm  is the mean square error (MSE) between the network output, $z_p$, and the true label, $t_p$, given by  
\begin{equation}\mathcal{L}=\frac{1}{2}\sum_{p=1}^S (z_p-t_p)^2,
\label{lossFL}
\end{equation}
where $t_p$ denotes the desired or target output (also called a teaching signal). 

\textbf{Training process in the convolutional layer.} To define the gradient descent relations for the update of all previous weights (within both the graph convolutional layer, $w_k(\nu)$, and   the fully connected layer, $v_p(m)$) in the training process, 
consider first the convolutional layer, described by (\ref{ConvCNNK})-(\ref{ConvCNN)}), to give the gradient weight update in the form
\begin{equation}w_{k}(\nu)_{new}= w_{k}(\nu)_{old}-\alpha \frac{\partial \mathcal{L}}{\partial w_{k}(\nu)} \ _{\Big|w_{k}(\nu)=w_{k}(\nu)_{old}}\label{GradUpCNN2L}\end{equation}
where $\alpha$ is a constant known as the step-size or learning rate. 
The element-wise gradient values in the first (convolutional) layer are designated by the superscript $(\cdot)^{(1)}$, and calculated as
\begin{equation}g_k^{(1)}(0)=\frac{\partial \mathcal{L}}{\partial w_{k}(0)}=\sum_n \frac{\partial \mathcal{L}}{{\color{red}\partial y_{k}(n)}}\frac{{\color{red}\partial y_{k}(n)}}{\partial w_{k}(0)}=\sum_n \frac{\partial \mathcal{L}}{\partial y_{k}(n)}x(n), \label{GradUpCNNGV}\end{equation}
\begin{gather}g^{(1)}(1)=\frac{\partial \mathcal{L}}{\partial w_{k}(1)}=\sum_n \frac{\partial \mathcal{L}}{{\color{red}\partial y_{k}(n)}}\frac{{\color{red}\partial y_{k}(n)}}{\partial w_{k}(1)} \nonumber \\ =\sum_n \frac{\partial \mathcal{L}}{\partial y_{k}(n)}\sum_{\mu}W_N(n,{\mu})x(\mu), \label{GradUpCNNGV2}\end{gather}
where the input-output relation in (\ref{ConvCNNK}) is used.

Next,  the terms $\partial \mathcal{L}/\partial y_{k}(n)$, called the \textit{delta error} function, $\Delta^{(1)}_k(n)$, are calculated using the chain rule, as
\begin{gather}\Delta^{(1)}_k(n)=\frac{\partial \mathcal{L}}{\partial y_{k}(n)}=\sum_p \frac{\partial \mathcal{L}}{{\color{red} \partial z_{p}}}\frac{{\color{red} \partial  z_{p}}}{\partial y_{k}(n)}=\sum_p \frac{\partial \mathcal{L}}{\partial z_{p}}\frac{\partial  z_{p}}{{\color{red}\partial o_{k}(n)}}\frac{{\color{red}\partial  o_{k}(n)}}{\partial y_{k}(n)} \nonumber \\
=\sum_p \Delta_p^{(2)}v_{p}((k-1)N+n-1) \ u(y_k(n)) \label{GradUpCNNGBP11}
\end{gather}
where $\partial  o_{k}(n)/\partial y_{k}(n)=u(y_k(n))$ according to (\ref{ConvCNN)}) and (\ref{ReLUdef}), while the relation in (\ref{ConvCNNFC}) is used for the calculation of $\partial  z_{p}/\partial o_{k}(n)=v_{p}((k-1)N+n-1)$, with 
$$\Delta_p^{(2)}=\frac{\partial \mathcal{L}}{\partial z_{p}}= z_p-t_p, \ \ \ p=1,2,\dots,S$$ as the delta error in the final (the second, in this case) stage. 

The relation in (\ref{GradUpCNNGBP11}) back-propagates the error from layer 2, denoted by $\Delta_p^{(2)}$, to layer 1, to give the portion of the overall error attributed to neuron $k$ of layer 1, denoted by $\Delta_k^{(1)}(n)$. We can now calculate $\partial \mathcal{L}/\partial y_{k}(n)=\Delta_k^{(1)}(n)$ and the gradient for the update in (\ref{GradUpCNN2L}). The gradient values of $g_k^{(1)}(0)$ and $g_k^{(1)}(1)$ in (\ref{GradUpCNNGV}) and (\ref{GradUpCNNGV2}) for the convolutional weight, $w_k(\nu)$, update can now be expressed as
\begin{gather}g_k^{(1)}(0)=\sum_n \frac{\partial \mathcal{L}}{\partial y_{k}(n)}x(n)=\sum_n \Delta^{(1)}_k(n)x(n), \text{\ \  and} \nonumber \\ 
g_k^{(1)}(1)=\sum_n \Delta^{(1)}_k(n)\sum_{\mu}W_N(n,{\mu})x(\mu),	
	\label{GradUpCNNGV9}\end{gather}
or in a matrix form
\begin{gather}\mathbf{g}_k^{(1)}=[\mathbf{x}^T \boldsymbol{\Delta}_k^{(1)},  \ \ (\mathbf{W}_N\mathbf{x})^T \boldsymbol{\Delta}_k^{(1)} ]^T=\begin{bmatrix} \mathbf{x}^T \\(\mathbf{W}_N\mathbf{x})^T\end{bmatrix}\boldsymbol{\Delta}_k^{(1)}.	
	\label{GradUpCNNGV9M}\end{gather}
This expression can easily be generalized for higher order  matched filters. For example, for $M=3$ and based on (\ref{ConvCNN3}), the gradient vector  would be of the form 
\begin{gather}\mathbf{g}_k^{(1)}=[\mathbf{x}^T \boldsymbol{\Delta}_k^{(1)},  \ \ (\mathbf{W}_N\mathbf{x})^T \boldsymbol{\Delta}_k^{(1)},  \ \ (\mathbf{W}^2_N\mathbf{x})^T \boldsymbol{\Delta}_k^{(1)} ]^T.	
	\label{GradUpCNNGV9M3}\end{gather}
The bias terms are updated in the same way as the weights in (\ref{GradUpCNN2L}), that is based on
\begin{equation}b_{k,new}=b_{k,old}-\beta \frac{\partial \mathcal{L}}
{\partial b_k} \ _{\Big|b_k=b_{k,old}} \label{BGradUpCNN22}
\end{equation}
and
\begin{equation}\frac{\partial \mathcal{L}}{\partial b_{k}}=\sum_n \frac{\partial \mathcal{L}}{{\color{red}\partial y_{k}(n)}}\frac{{\color{red}\partial y_{k}(n)}}{\partial b_{k}}=\sum_n \frac{\partial \mathcal{L}}{\partial y_{k}(n)}=\sum_n \Delta^{(1)}_k(n). \label{GradUpCNNB}\end{equation}

\noindent\textbf{ Fully Connected (FC) layer.}  The input to the FC layer represents the flattened output from the convolutional layer, given by
$$o_F((k-1)N+n)=o_F(m),$$
where the indices $m$ in $o_F(m)$ range from $1$ to $KN$, with $k=1,2,\dots,K$ and $n=1,2,\dots,N$. 
Notice that the relation (\ref{ConvCNNFC}) can be equally written as 
$$z_p=\sum_{m=1}^{KN}v_{p}(m-1)o_F(n).$$

The update of the fully connected layer weights, $v_{p}(m)$, is then performed in the same way as in (\ref{GradUpCNN2L}), using
\begin{equation}v_{p}(m)_{new}= v_{p}(m)_{old}-\gamma \frac{\partial \mathcal{L}}{\partial v_{p}(m)} \ _{{{\Big|v_{p}(m)=v_{p}(m)_{old}}}},\label{GradUpCNN2FC}\end{equation}
with the gradient elements in the form
$$g^{(2)}_p\!\!(m)\!=\!\!\frac{\partial \mathcal{L}}{\partial v_{p}(m)}\! = \!\! \frac{\partial \mathcal{L}}{{\color{red}\partial z_{p}}} \frac{{\color{red}\partial z_{p}}}{\partial v_{p}(m)}\!\!=\!(z_p\!-\!t_p)o_F(m)\!=\!\Delta_p^{(2)}\!\!o_F(m),$$
and $\gamma$ is the step-size.

If a nonlinear activation function is used at the output,  the factor of $f'(z_p)$ will correspondingly multiply the right hand side of $\partial \mathcal{L}/\partial v_{p}(m)$.

\subsubsection{ Softmax Output Layer} 

In some applications, the output layer should give the probabilities for the decision when classifying of the analyzed data. Such an output therefore represents the probabilities for different possible labels (basins of attraction) associated with the analyzed signal or image (for example, dog, cat, bird in the image), whereby  the label that receives the highest probability is the overall classification decision. In the error calculation, the desired (target) output then assumes the value $t_p=1$ for one value  $p=p_0$ (in the training process, we know what signal/image is analyzed by the GCNN) and $t_p=0$ for other values of $p=1,2,\dots,S, p \ne p_o$.

Since the output, $z_p$, from the last layer (overall output), may assume various positive and negative real values, we need to map the output $z_p$ onto probability-like values.  This is achieved using a function of the form
\begin{equation}P_p=\frac{e^{z_p}}{\sum_{i=1}^Se^{z_i}}, \ \ \ p=1,2,\dots,S.\label{softmaxM}\end{equation}
called the softmax. Obviously, $0 \le P_p \le 1$ and $\sum_{p=1}^SP_p=1$. 

When the softmax is used as the output mapping, the loss function is modified accordingly, that is, from the mean square error to the 
 cross-entropy form, given by
$$\mathcal{L}=-\sum_{p=1}^St_p\ln(P_p).$$
Physically, cross-entropy is very large if there is a target $t_p$ close to $1$, but the corresponding output probability $P_p$ is small, indicating that a big change in the weights. Conversely, the cross-entropy, $\mathcal{L}$, is small only when for $t_{p_0}=1$ at a specific $p_0$, and the value of corresponding $P_{p_0}$ is close to $1$. 

We can easily show that the delta error function in the  output layer is of the form 
$$\Delta^{(2)}_p=\frac{\partial \mathcal{L}}{\partial z_p}=\sum_{i=1}^S\frac{\partial \mathcal{L}}{{\color{red}\partial P_i}}\frac{{\color{red}\partial P_i}}{\partial z_p}=\sum_{i=1}^S\Big(\frac{t_i}{P_i}P_iP_p\Big)-\frac{t_p}{P_p}P_p=P_p-t_p$$
since from (\ref{softmaxM}) it follows that $\partial P_i /\partial z_p=-P_iP_p$ if $i \ne p$  and $\partial P_i /\partial z_p=P_i(1-P_p)=-P_iP_p+P_p$ if $i=p$, while $\sum_{i=1}^St_i=1$. 

Therefore, as expected, there is no weight correction if $t_p=P_p$, while, as desired, all other  relations regarding the back-propagation hold in this case, without any modification.

\begin{figure}[ptb]
	\begin{center}
		\includegraphics[scale=0.85]{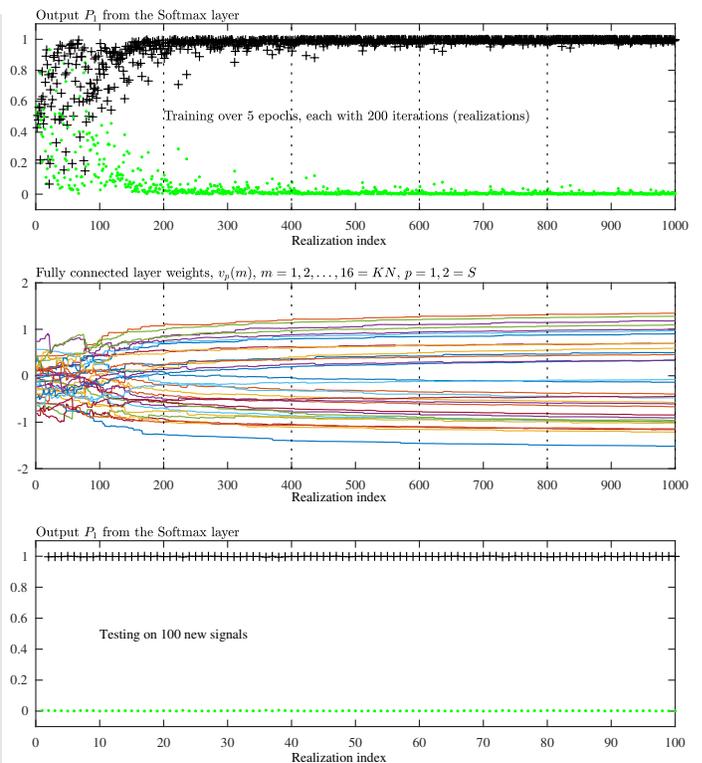} 
		\caption{Operation of a GCNN with one graph convolution layer and one FC layer, with two neurons at the output (softmax) layer and $K=2$ channels.  The FC layer  therefore had $(NK) \times 2 = 16\times2=32$ weights.
			(Top) The obtained probability, $P_1$, at the output of the GCNN (the first output of the softmax layer) is designated using black "+" for the values when the correct result should be $P_1=1$. The mark is "+" is used to show the value of $P_1$ when the $\mathbf{feature}_1$ is present. The output values $P_1$ are designated by green "$\cdot$" when the correct output value should be $P_1=0$, that is when $\mathbf{feature}_2$ is present in the input signal.
			(Middle) The evaluation of weights in the FC layer during the training process. (Bottom) The  test results of the probability at the first output of the softmax layer, $P_1$, after the training process, with the marks "+" and "$\cdot$" as in the top panel.}%
		\label{Example_CNN}%
	\end{center}
	
\end{figure}

\begin{exmpl}\label{Example3} To illustrate the operation of the  GCNN we consider a simple two-layer neural network (one graph convolutional layer and one fully connected layer), as shown in Fig. \ref{fig-sig-gr1-CNN} and evaluate all calculations step-by-step. 
	
	We consider an input noisy signal with $N=8$ samples, which  contains either a variant of the  first feature, 
 \begin{gather}
	\mathbf{x}=\mathbf{feature}_1=\mathbf{x}_0-\mathbf{D}^{-1/2}\mathbf{W}\mathbf{D}^{-1/2}\mathbf{x}_0 \label{feature1}
\end{gather} 	
	 or a variant of the second feature, 
	 \begin{gather}
	 	\mathbf{x}=\mathbf{feature}_2=\mathbf{x}_0+\mathbf{D}^{-1/2}\mathbf{W}\mathbf{D}^{-1/2}\mathbf{x}_0
	 \end{gather} 
 where the assumed graph signal, $\mathbf{x}_0$, has the form $x_0(n)=\delta(n-n_0)$ with a random $n_0$. The target signal for $\mathbf{feature}_1$ is $t=[1, \ 0]$ and $t=[0, \ 1]$ for $\mathbf{feature}_2$. The weight matrix, $\mathbf{W}$, is given by (\ref{matA1a}). The considered features are similar to the graph signals shown in Fig. \ref{fig-sig-gr1}(c),(d).
 
 Graph convolutional filters of length $M=2$, defined by (\ref{eq:filter-timeCNN2222}), were used to produce $K=2$ channels at the graph convolutional layer. The softmax was used at the output of the FC layer, with two values that correspond to the two patterns in the target signal, $\mathbf{t}$. 
 
 \begin{itemize}
 	\item Training was performed based on $200$ random realizations of the input signal, randomly assuming $\mathbf{feature}_1$ or $\mathbf{feature}_2$ and with a random central vertex, $n_0$, for each realization. This cycle of $200$ realizations is called an epoch. Then, the same set of $200$ random realizations was repeated  $10$ times (10 epochs were used in training), that is, the GCNN  was trained over 10 epochs, with no max-pooling  used.

	\item  Fig. \ref{Example_CNN} illustrates the training process and the values of the corresponding parameters.

	The obtained probability, $P_1$, at the first output of the GCNN (the first output of the softmax layer) is given in the top panel of Fig. \ref{Example_CNN}. The black "+" designate the values when the correct result should be $P_1=1$, that is, the symbol "+" shows the obtained value of $P_1$ when the $\mathbf{feature}_1$ is present in the input noisy signal. The output values $P_1$ are designated by the green "$\cdot$" when the correct output value should be $P_1=0$, that is, when $\mathbf{feature}_2$ is present in the input signal. Note that the other output of the softmax layer, denoted by $P_2$, is such that $P_1+P_2=1$ always holds. 
	
	In an ideal case all black "+" marks should be in positions where the probability (GCNN softmax output) value is equal to 1, while all green "."  marks should be in the positions where the value of $P_1$ is equal to 0. 
	
From the top panel of Fig. \ref{Example_CNN}, 	we can see that the output (softmax) probability, $P_1$, initially assumes indecisive values around 0.5 and then converges to the ideal ones (either 1 or 0) during the training process (over the iterations and epochs). 
	
	\item The values of weights in the FC layer during the training process, are given in the middle panel in Fig. \ref{Example_CNN}, illustrate evaluation of the weight adaptation.

	\item After the GCNN is trained over 10 epochs of 200 random realizations of the signal, the update process of the weights in all layers is stopped, and the so obtained weights are tested over 100 new random realizations of the input. Again the output probability $P_1$ was considered. The results are shown in the bottom panel of Fig. \ref{Example_CNN}. Observe that the decision was correct and highly reliable in all 100 new cases, with the marks black  "+" and green "$\cdot.$"  used in the same way as described in the second item of this list. 
	
	\item  A step-by-step calculation in the first iteration of the training process is given in Appendix B.

\end{itemize}
\end{exmpl}

\section*{Appendix A: Max-Pooling through Graph coarsening}

Graph coarsening belongs to graph down-sampling strategies and refers to the reduction in the number of vertices of the original graph \cite{stankovic2020data1,stankovic2020data3}. Graph coarsening is typically used in graph partitioning and for the visualization of large graphs in a computationally efficient manner \cite{tremblay2020approximating}. In general, it can be performed by grouping the vertices into $N_c<N$ nonoverlapping groups, subsequently forming new vertices, and finally connecting these new vertices (former groups of vertices) with the equivalent weights, which represent a sum of all weights between the groups. The weight matrix of the so coarsened graph is given by
$$\mathbf{W}_c=\mathbf{P}\mathbf{W}\mathbf{P}^T,$$ 
where $\mathbf{P}$ is the indicator matrix of the groups of vertices \cite{stankovic2020data1}. 

For the max-pooling operation we used the first iteration of the graph signal in the first channel of the considered GCNN, after the ReLU operation, given by $f(\mathbf{y}_1)=\begin{bmatrix}
	0.012 & 0.037 & 0 &  0.121 & 0 & 0 & 0 & 0.053  \end{bmatrix}^T$ (see Appendix B).  Observe from  Fig. \ref{Example_CNN_coars} (a) that the maximum signal  value  is at the vertex $n=4$. We should therefore avoid all signal values within the one-neighborhood of the vertex $n=4$. This means that the vertices $n=2,3,5,6$ are fused (merged) with vertex $n=4$ to form the super-vertex 
{\footnotesize $23$}$\textbf{4}${\footnotesize $56$}, with the associate signal value $f(y_1(4))=0.121$, as shown in Fig. \ref{Example_CNN_coars}(b). The generation of this super-vertex in (\ref{matA1P}) is defined by the second row of the indicator matrix, $\mathbf{P}$. After these vertices, together with the corresponding signal values, are excluded, the remaining maximum signal value is at the vertex $n=8$. This vertex is then fused with its neighboring vertices at the distance of one, excluding the already formed super-vertex. Then, a new super-vertex, denoted by  {\footnotesize $1$}$\textbf{8}$ is formed. The row in the indicator matrix, $\mathbf{P}$, corresponding to this super-vertex is denoted by {\footnotesize $1$}$\textbf{8}$. Finally, the remaining vertex is the vertex $n=7$. When the indicator matrix $\mathbf{P}$ is formed, the weight matrix, $\mathbf{W}_c=\mathbf{P}\mathbf{W}\mathbf{P}^T$ in (\ref{matA1PP}), is calculated for the three new vertices {\footnotesize $1$}$\textbf{8}$, {\footnotesize $23$}$\textbf{4}${\footnotesize $56$}. Fig. \ref{Example_CNN_coars}(b) shows such a  coarsened graph which corresponds to max-pooling. Note that this graph topology changes for each channel and each iteration, since it is signal dependent.  The corresponding matrices are given below.  

\begin{gather}
	\!\!\mathbf{W}=
	\begin{array}{cr}
		& \\
		{
			\color{blue}
			\begin{matrix}
				\text{\footnotesize 1}\\
				\text{\footnotesize 2}\\
				\text{\footnotesize 3}\\
				\text{\footnotesize 4}\\
				\text{\footnotesize 5}\\
				\text{\footnotesize 6}\\
				\text{\footnotesize 7}\\
				\text{\footnotesize 8}\\
			\end{matrix}
		} &
		\begin{bmatrix}
			\ 0\  & \ 1\  & \ 1\  & \ 0\  & \ 0\  & \ 0\  & \ 0\  & \ 1\  \\
			\ 1\  & \ 0\  & \ 1\  & \ 1\  & \ 1\  & \ 0\  & \ 0\  & \ 1\  \\
			\ 1\  & \ 1\  & \ 0\  & \ 1\  & \ 1\  & \ 0\  & \ 0\  & \ 0\  \\
			\ 0\  & \ 1\  & \ 1\  & \ 0\  & \ 1\  & \ 1\  & \ 0\  & \ 0\  \\
			\ 0\  & \ 1\  & \ 0  & \ 1\  & \ 0\  & \ 1\  & \ 1\  & \ 1\  \\
			\ 0\  & \ 0\  & \ 0\  & \ 1\  & \ 1\  & \ 0\  & \ 1\  & \ 0\  \\
			\ 0\  & \ 0\  & \ 0\  & \ 0\  & \ 1\  & \ 1\  & \ 0\  & \ 0\  \\
			\ 1\  & \ 1\  & \ 0\  & \ 0\  & \ 1\  & \ 0\  & \ 0\  & \ 0\ 
		\end{bmatrix} \\
		& 
		{
			\color{blue}
			\begin{matrix}
			  \text{\footnotesize 1}  &
				\ \text{\footnotesize 2\ }\  &
				\ \text{\footnotesize 3}\  &
				\ \text{\footnotesize 4}\  &
				\ \text{\footnotesize 5}\  &
				\ \text{\footnotesize 6\ }\  &
				\ \text{\footnotesize 7}\  &
				 \text{\footnotesize 8} &
			\end{matrix}
		}
	\end{array}\!\!, \label{matA1a}
\end{gather}

\begin{gather}
	\!\!\mathbf{P}=
	\begin{array}{cr}
		& \\
		{
			\color{blue}
			\begin{matrix}
				\text{\footnotesize 1,8}\\
				\text{\footnotesize 2,3,4,5,6}\\
				\text{\footnotesize 7}\\
			\end{matrix}
		} &
		\begin{bmatrix}
			\ 1\  & \ 0\  & \ 0\  & \ 0\  & \ 0\  & \ 0\  & \ 0\  & \ 1\  \\
			\ 0\  & \ 1\  & \ 1\  & \ 1\  & \ 1\  & \ 1\  & \ 0\  & \ 0\  \\
			\ 0\  & \ 0\  & \ 0\  & \ 0\  & \ 0\  & \ 0\  & \ 1\  & \ 0\ 
		\end{bmatrix} \\
		& 
		{
			\color{blue}
			\begin{matrix}
			\!\!\!	\text{\footnotesize 1}  &
				\ \text{\footnotesize 2\ }\  &
				\ \text{\footnotesize 3}\  &
				\ \text{\footnotesize 4}\  &
				\ \text{\footnotesize 5}\  &
				\ \text{\footnotesize 6\ }\  &
				\ \text{\footnotesize 7}\  &
				\ \text{\footnotesize 8}\!\! &
			\end{matrix}
		}
	\end{array}\!\!, \label{matA1P}
\end{gather}

\begin{gather}
\!\!\mathbf{W}_c=\mathbf{P}\mathbf{W}\mathbf{P}^T=
\begin{array}{cr}
	& \\
	{
		\color{blue}
		\begin{matrix}
			\text{\footnotesize 1,8}\\
			\text{\footnotesize 2,3,4,5,6}\\
			\text{\footnotesize 7}\\
		\end{matrix}
	} &
	\begin{bmatrix}
		\ 2\  & \ 4\  & \ 0\   \\
		\ 4\  & \ 14\  & \ 2\    \\
		\ 0\  & \ 2\  & \ 0\
	\end{bmatrix} \\
	& 
	{
		\color{blue}
		\begin{matrix}
			\text{\tiny 1,8} \!\!\!\!\!\!  &
			\ \text{\tiny 2,3,4,5,6\ } &
			\text{\tiny 7}\!\!\! &
		\end{matrix}
	}
\end{array}\!\!, \label{matA1PP}
\end{gather}

\noindent\textbf{The indicator matrix.}	A matrix which indicates the values of the signal that kept their nonzero values after the ReLU operation and the max-pooling, is given by  \\
\begin{gather}
\mathbf{M}^{ReLU+MP}=\arraycolsep3pt\begin{bmatrix}
	0 & 0 & 0 & \textbf{1} & 0 & 0   &  \textbf{0} &\textbf{1}  \\
	0 &  \textbf{1} &0  & 0  &0  & 0  &\textbf{1}&0
\end{bmatrix}^T.\label{MpMatix}\end{gather}

The first row shows that the signal values at the vertices $n=4$ and $n=8$ “survived" both operations, the ReLU operation and the max-pooling, in the first channel. In a similar way, the signal values at the vertices, $n=7$ and $n=2$  “survived" in the second channel.  This matrix would be used to reposition the gradient updates to the proper positions if the max-pooling were used.

\begin{figure}[hptb]
	\begin{center}
		\includegraphics{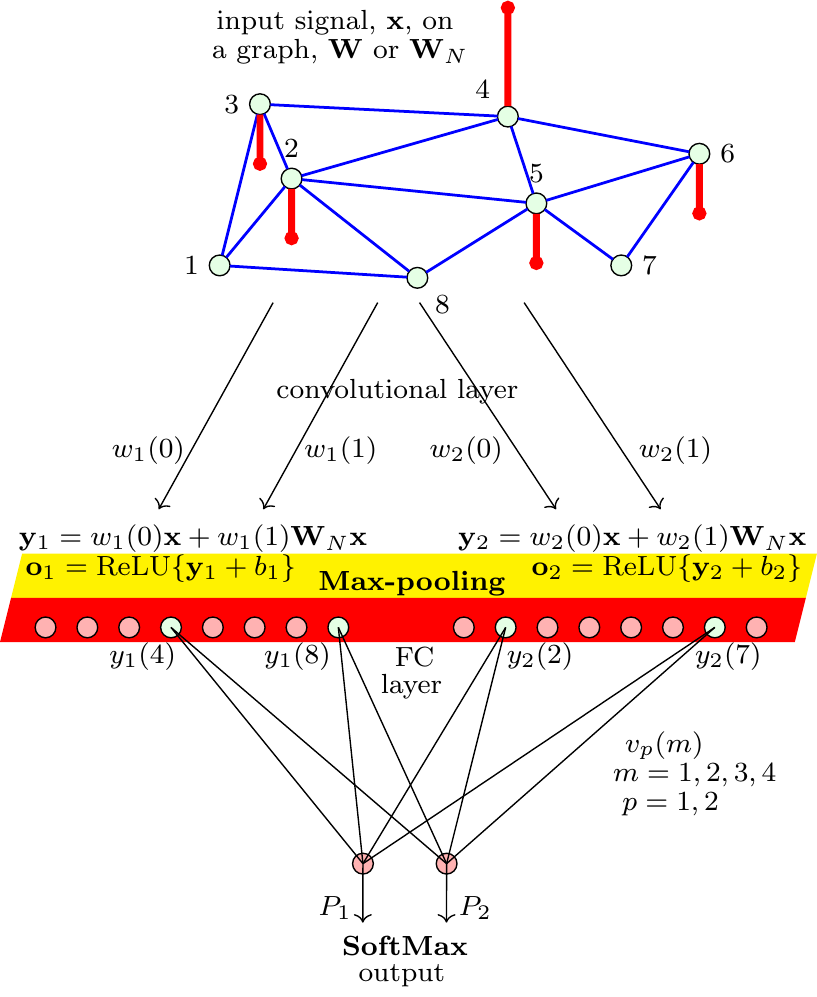} 
		\caption{Principle of a GCNN with max-pooling (which corresponds to the signal values selected by the indicator matrix in (\ref{MpMatix})). The operation is illustrated based on a GCNN with one graph convolution layer and one FC layer, and  two neurons at the output (softmax) layer. }%
		\label{fig-sig-gr1-CNN_MP}%
	\end{center}
\end{figure}

The GCNN from Fig. \ref{fig-sig-gr1-CNN}, with max-pooling included, is shown in Fig. \ref{fig-sig-gr1-CNN_MP}. For more detail on the implementation, please see our sister paper \cite{CNN_Tutorial}.

\noindent\textbf{Graph lifting (uncoarsening).} Graph  lifting is an inverse operation to graph coarsening, and represents a process of obtaining a larger scale (fine) graph from a coarsened (smaller) graph. The weight matrix, $\mathbf{W}_L$, of the lifted graph is obtained from the weight matrix of the coarsened graph, $\mathbf{W}_c$, as
$$\mathbf{W}_L=\mathbf{P}^+\mathbf{W}_c(\mathbf{P}^+)^T,$$
where $\mathbf{P}^+$ is the pseudo-inverse of the indicator matrix, such that $\mathbf{P}\mathbf{P}^+=\mathbf{I}$, where  $\mathbf{I}$ is the identity matrix. 

\begin{figure}[hptb]
	\begin{center}
			\includegraphics[scale=1]{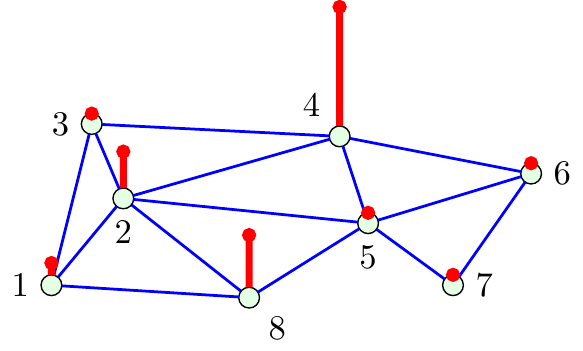} 
			(a)
			
		\includegraphics[scale=1]{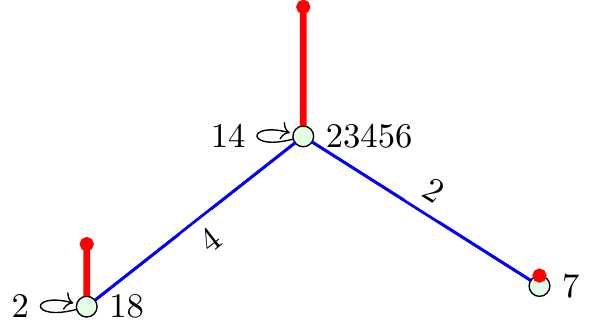} 	
		(b)
		
		\caption{Illustration of the max pooling operation in a graph CNN implemented using graph coarsening. (a) An example of a graph signal. (b) The coarsened graph using the max-polling of the graph signal value at the current vertex, within its neighborhood one.}%
		\label{Example_CNN_coars}%
	\end{center}
	
\end{figure}

The same relations as for the weights hold for the corresponding graph Laplacian of the original graph, $\mathbf{L}$, graph Laplacian of the coarsened graph, $\mathbf{L}_c$, and the graph Laplacian of the lifted graph, $\mathbf{L}_L$, that is
\begin{gather*}\mathbf{L}_c=\mathbf{P}\mathbf{L}\mathbf{P}^T.\\
	\mathbf{L}_L=\mathbf{P}^+\mathbf{L}_c(\mathbf{P}^+)^T.
\end{gather*}

Notice that for the normalized graph Laplacian, the definition of the indicator matrix should be slightly modified \cite{jin2020graph}.

\noindent\textbf{ Generalization of graph coarsening.} The process of  coarsening a graph, $\mathcal{G}$ (with vertices $\mathcal{V}$, edges $\mathcal{B}$, and weights $\mathcal{W}$),  may be continued until a desired number of vertices is obtained. 
In general, the coarsening involves a sequence of graphs 
\begin{gather*}
	\mathcal{G}=\mathcal{G}_0=\{\mathcal{V},\mathcal{B},\mathcal{W}\}=\{\mathcal{V}_0,\mathcal{B}_0,\mathcal{W}_0\} \\
	\mathcal{G}_1=\{\mathcal{V}_1,\mathcal{B}_1,\mathcal{W}_1\} \\
	\vdots \\
	\mathcal{G}_c=\{\mathcal{V}_c,\mathcal{B}_c,\mathcal{W}_c\},
\end{gather*}
whereby at every iteration, the coarsened graph, $\mathcal{G}_{l+1}=\{\mathcal{V}_{l+1},\mathcal{B}_{l+1},\mathcal{W}_{l+1}\}$, 
is obtained from the previous one through a weight matrix transformation based on the corresponding indicator matrices,
$$\mathbf{W}_{l}=\mathbf{P}_l\mathbf{W}_{l-1}\mathbf{P}_l^T,$$
while the lifting is performed as $\mathbf{W}_{l-1}=\mathbf{P}_l^+\mathbf{W}_{l}(\mathbf{P}_l^+)^T$.

\bibliographystyle{ieeetr}

\bibliography{graph-signal-processing}

\onecolumn

\section*{Appendix B: A step-by-step calculation in the first iteration of the GCNN from Example \ref{Example3}}

\footnotesize

\begin{tabular}{|l|l|}
	
	\hline 
	
		\textbf{Forward calculation}: From the input signal to the output  \\
		\hline
		$\bullet$\textbf{ Input} signal, $\mathbf{x}$, of length $N=8$, 
		\\
		$\mathbf{x}=\begin{bmatrix}
			0.087 & 0.030 & -0.006 & 0.039 & -0.254 & -0.426 & 0.946 & -0.145 \end{bmatrix}^T.$ \\ The target signal was $\mathbf{t}=\begin{bmatrix}1 & 0\end{bmatrix}^T$, since $\mathbf{feature}_1=\mathbf{x}_0-\mathbf{W}_N\mathbf{x}_0=[ 0,         0,         0,         0,  -0.316,   -0.408,  \  1,  \        0]^T$ \\  was present in the input (with a small noise). This feature was obtained using (\ref{feature1}) and $x_0(n)=\delta(n-7)$. \\
		 The normalized weight matrix, $\mathbf{W}_N$ is given by (\ref{matA1aL}),
		\\
		\hline
		$\bullet$\textbf{ Weight initialization}: Random $w_k(\nu) \sim \mathcal{N}(0,1)\sqrt{2/M}$, $M=2$, for $K=2$ channels, 	$\mathbf{w}_k=[w_k(0), \ w_k(1)]^T$,: \\
		$\mathbf{w}_1=\begin{bmatrix}
			-0.221 & -0.741 \end{bmatrix}^T$,\\
		$\mathbf{w}_2=\begin{bmatrix}
			\phantom{-}1.429 & \phantom{-}0.323 \end{bmatrix}^T$,    \\
		\hline
		$\bullet$\textbf{ Convolutions}:  $\mathbf{y}_k=\mathbf{w}_k*_c\mathbf{x}+b_k=w_k(0)\mathbf{x}+w_k(1)\mathbf{W}_N\mathbf{x}, \ \  \ k=1,2$, \\  with the initial bias values $b_k=0$ and $k=1,2$. \\ 
		$\mathbf{y}_1=\begin{bmatrix}
			\phantom{-}0.012 & \phantom{-}0.037 & -0.034 & \phantom{-}0.121 & -0.067 & -0.152 & -0.021 & \phantom{-}0.053  \end{bmatrix}^T
		$, \\
		$\mathbf{y}_2=\begin{bmatrix}
			\phantom{-}0.111 & \phantom{-}0.024 & \phantom{-}0.007 & -0.001 & -0.309 & -0.501 & \phantom{-}1.270 & -0.217 \end{bmatrix}^T
		$.
		\\
		\hline
		$\bullet$\textbf{ Nonlinear activation function}:  \textbf{ ReLU} activation function, $
		f(\mathbf{y}_k)=\max\{0,\mathbf{y}_k\}$, was used, to give \\	
		$f(\mathbf{y}_1)=\begin{bmatrix}
		0.012 & 0.037 & {\color{red}0} &  0.121 & {\color{red}0} & {\color{red}0} & {\color{red}0} & 0.053  \end{bmatrix}^T
		$ 
		\\
		$f(\mathbf{y}_2)=\begin{bmatrix}
			0.011 & 0.024 & 0.007 & {\color{red}0} & {\color{red}0} &  {\color{red}0} & 1.270 & {\color{red}0} \end{bmatrix}^T$,   \\
		\hline
		$\bullet$\textbf{ Indicator matrix}: 
		The indicator matrix of chosen (nonzero) values from the ReLU, $\mathbf{M}^{ReLU}$, \\
		$\mathbf{M}^{ReLU}=\arraycolsep3pt\begin{bmatrix}
			 \textbf{1} &\textbf{1} & {\color{red} 0} & \textbf{1} & {\color{red} 0} & {\color{red} 0}   &  {\color{red} 0} &\textbf{1}  \\
			\textbf{1}  &  \textbf{1} &\textbf{1} &  {\color{red} 0}  & {\color{red} 0}  & {\color{red} 0}  &\textbf{1}& {\color{red} 0}
		\end{bmatrix}^T$.  \\
		It would be used to reposition the gradient update to the proper positions if the downsampled signal was used, \\ taking into account the zeroing by the ReLU.  \\
		\hline
		$\bullet$\textbf{ Flattening}: $o_F((k-1)N+n)=f(y_k(n))$, \ \ $k=1,2, \ \ n=1,2,3,4,5,6,7,8$. \\
		$\mathbf{o}_F=\begin{bmatrix} 0.012, \ 0.037, {\color{red} \ 0}, \  0.121, {\color{red} \ 0, \ 0, \ 0}, \ 0.053, \ 0.011, {\color{red} \ 0}, \ 0.024, \ 0.007, {\color{red} \ 0, \ 0, \ 0}, \ 1.270, {\color{red} \ 0} \end{bmatrix} ^T$, \\
		\hline
		$\bullet$\textbf{ Weight initialization}: For the FC layer, random weights $v_p(m) \sim \mathcal{N}(0,1)\sqrt{2/(NK)}=\mathcal{N}(0,1)\sqrt{1/8}$, \\
		$\mathbf{v}=\begin{bmatrix}-0.045, \phantom{-}0.391,  -0.289, \phantom{-}0.123, \phantom{-}0.309, \phantom{-}0.029, \phantom{-}0.121, -0.132,  -0.389,  \phantom{-}0.081,  -0.055,  -0.609,  -0.183,  -0.765,  \phantom{-}0.277,  \phantom{-}0.174 \\
		-0.248 ,  \phantom{-}0.023,   -0.085,  \phantom{-}0.543,  \phantom{-}0.102,  -0.548 , -0.542, -0.360, \phantom{-}0.706,  \phantom{-}0.412,  -0.590,  -0.714, -0.445,  \phantom{-}0.102,  \phantom{-}0.245,  -0.226
		\end{bmatrix}^T$ \\
		\hline
		$\bullet$\textbf{ Output}: The FC layer output signal, $z_p=\sum_{m=1}^{16} o_F(m)v_p(m-1)$. \\
		$\mathbf{z}= [z_1, \ z_2]^T=\mathbf{v}^T\mathbf{o}_F=\begin{bmatrix} 0.332 & 0.440
		\end{bmatrix}^T$.  \\
		\hline
		$\bullet$\textbf{ Softmax}: With $S=2$ output values,  \  $P_p=e^{z_p}/(e^{z_1}+e^{z_2})$, \ $p=1,2$, we get \\
		$\mathbf{P}=[P_1, \ \ P_2]=\begin{bmatrix}
		0.4731 &   0.5269 \end{bmatrix}^T$.  \ \ \ 
	 \\
		\hline
	\end{tabular}
	
\bigskip

\begin{tabular}{|l|l|}
	
	\hline 
	\textbf{Back-propagation}: Delta error, gradient, weight updates \\
	\hline 
	$\bullet$\textbf{ Output Delta error}
		$\boldsymbol{\Delta}^{(2)}=[\Delta^{(2)}_1, \ \Delta^{(2)}_2]^T=\mathbf{P}-\mathbf{t}=\begin{bmatrix}
		-0.5269  &  0.5269\end{bmatrix}^T$, \ \ $\Delta^{(2)}_p=P_p-t_p$\\
	\hline
	$\bullet$\textbf{ Gradient} for the FC layer weights update, $g^{(2)}_p(m)=\Delta^{(2)}_p o_F(m)$; \ \ $o_F(m)$ is the input to the FC layer and $\Delta^{(2)}_p$ is the output Delta error\\
	$\mathbf{g}^{(2)}=\mathbf{o}_F(\mathbf{\Delta}^{(2)})^T=\begin{bmatrix}
	-0.006, -0.019,  \ 0, -0.064, \ 0, \ 0, \ 0, -0.028, -0.058, -0.013, -0.004, \ 0, \ 0, \ 0, -0.669, \ 0 \\
	\phantom{-}0.006, \phantom{-}0.019, \ 0, \phantom{-}0.064, \ 0, \ 0, \ 0, \phantom{-}0.028, \phantom{-}0.058, \phantom{-}0.013, \phantom{-}0.004, \ 0, \ 0, \ 0, \phantom{-}0.669, \ 0
	\end{bmatrix}^T$,  \\
	\hline
	$\bullet$\textbf{ Weight update} in the FC layer using the gradient $\mathbf{g}^{(2)}$ and the  step $\alpha=0.1$, \ \ $\mathbf{v}\leftarrow\mathbf{v}-0.1\mathbf{g}^{(2)}$,\\
	$\mathbf{v}=\begin{bmatrix}
		-0.044, \phantom{-} 0.394, {\color{red}-0.289}, \phantom{-} 0.132, {\color{red}\phantom{-} 0.309, \phantom{-}0.029, \phantom{-}0.121}, -0.128, -0.380, \phantom{-}0.082, -0.054, {\color{red}-0.609, -0.183, -0.765}, \phantom{-}0.378, {\color{red}\phantom{-}0.174} \\
		-0.249, \phantom{-}0.021, {\color{red}-0.085}, \phantom{-}0.534, {\color{red}\phantom{-}0.102, -0.548, -0.542}, -0.364, \phantom{-}0.697, \phantom{-}0.410, -0.591, {\color{red}-0.714, -0.445, \phantom{-}0.102}, \phantom{-}0.145, {\color{red}-0.226}
	\end{bmatrix}^T$ \\
The unchanged weights in this iteration (in red) are defined by the zero-valued input, $o_F(m)$, to the FC layer (concatenated matrix $\mathbf{M}^{ReLU}$)\\
	\hline
	$\bullet$\textbf{ Delta error back-propagation} from the output, $\Delta^{(2)}_p$, to the convolutional layer,  $\Delta^{(1)}_k(n)=\sum_p\Delta^{(2)}_pv_p((k-1)N+n-1) \ u(y_k(n))$,  \\
	$\boldsymbol{\Delta}_1^{(1)}=\begin{bmatrix}-0.108 & -0.197 & \ \ {\color{red} 0} & 0.211 & {\color{red} 0} & {\color{red} 0} &{\color{red} 0} & -0.125\end{bmatrix}^T$,    \\
	$\boldsymbol{\Delta}_2^{(1)}=\begin{bmatrix}\phantom{-}0.567 & \phantom{-}0.173 & -0.283 & {\color{red} 0} & {\color{red} 0} & {\color{red} 0} & -0.123 & {\color{red} 0} \end{bmatrix}^T.$  \ \ Notice that the elements $u(y_k(m))$ in (\ref{GradUpCNNGBP11}) is defined by $\mathbf{M}^{ReLU}$\\
	\hline
	$\bullet$\textbf{ Gradient} for the weight update in the convolutional layer, ˆ$\mathbf{g}_k^{(1)}=[\mathbf{x}^T \boldsymbol{\Delta}_k^{(1)},  \ \ (\mathbf{W}_N\mathbf{x})^T \boldsymbol{\Delta}_k^{(1)} ]^T$, $k=1,2$, \\
	$\mathbf{g}^{(1)}_1=\begin{bmatrix} \phantom{-}0.011 & -0.017 \end{bmatrix}^T$, \\ 
	$\mathbf{g}^{(1)}_2=\begin{bmatrix}-0.060 & -0.017\end{bmatrix}^T$,   \\
	\hline
	$\bullet$\textbf{ Weight update} in the convolutional layer $\mathbf{w}_k\leftarrow \mathbf{w}_k-0.1\mathbf{g}^{(1)}_k$, $k=1,2$, \\
	$\mathbf{w}_1=\begin{bmatrix}
		-0.223 & -0.739 \end{bmatrix}^T$ \\
	$\mathbf{w}_2=\begin{bmatrix}
		\phantom{-}1.437 & \phantom{-}0.325 \end{bmatrix}^T$, \\
	\hline
	$\bullet$\textbf{ Bias update}, According to (\ref{GradUpCNNB}), $b_k \leftarrow b_k-0.05\sum_n \Delta_k^{(1)}(n)$, $k=1,2$, \\
	$\mathbf{b}= \mathbf{0}-0.05(\begin{bmatrix}1&1&1&1&1&1&1&1\end{bmatrix} \boldsymbol{\Delta}^{(1)})^T = \begin{bmatrix}
		 0.0109, \   -0.0167
		   \end{bmatrix}^T$. \\
	\hline
	$\bullet$\textbf{ New iteration} with a new signal realization,  \\
	$\mathbf{x}=\begin{bmatrix}
		-0.412 & 0.886 & -0.338 & -0.202 & -0.204 & 0.029 & 0.035 & -0.304 \end{bmatrix}^T$, \ \  $\mathbf{t}=[1, \ \ 0]$, \\
	Go back to the first step with the new (updated) weights, $\mathbf{w}$ and $\mathbf{v}$, and bias $\mathbf{b}$. \\
	Some of the results (the softamax output signal, $P_1$, and the FC layer weights, $v_p(m)$) over 1000 iterations are  shown in Fig. \ref{Example_CNN}. \\
	\hline
\end{tabular}

\medskip

$n=1,2,\dots,N$ is the vertex index in the input layer 

$k=1,2,\dots,K$ is the channel (matched filter) index

$\nu=1,2,\dots,M$ is the matched filter order index

$m=1,2,\dots,KN$ is the FC input node index

$p=1,2,\dots,S$ is the output node index

\end{document}